\documentclass[10pt,journal,cspaper,compsoc]{IEEEtran}

\usepackage{times}
\usepackage{epsfig}
\usepackage{graphicx}
\usepackage{amsmath}
\usepackage{amssymb}
\usepackage{multirow}
\usepackage{color}
\usepackage[table]{xcolor}
\usepackage{overpic}
\usepackage{wrapfig,lipsum,booktabs}
\usepackage{url}
\usepackage[nocompress]{cite} 
\usepackage[caption=false]{subfig}
\usepackage{tabularx}
\usepackage{ragged2e}
\definecolor{ballblue}{rgb}{0.13, 0.67, 0.8}

\usepackage[colorlinks,bookmarksnumbered,bookmarksopen]{hyperref}
\usepackage{geometry}
\ifdefined \GramaCheck
  \newcommand{\CheckRmv}[1]{}
  \newcommand{\figref}[1]{Figure 1}%
  \newcommand{\tabref}[1]{Table 1}%
  \newcommand{\secref}[1]{Section 1}
  \renewcommand{\eqref}[1]{Equation 1}
\else
  \newcommand{\CheckRmv}[1]{#1}
  \newcommand{\figref}[1]{Fig.~\ref{#1}}%
  \newcommand{\tabref}[1]{Table~\ref{#1}}%
  \newcommand{\secref}[1]{Sec.~\ref{#1}}
  \renewcommand{\eqref}[1]{Equation~(\ref{#1})}
\fi

\newcommand{\cmm}[1]{{\textcolor{blue}{#1}}}

\newcommand{\myPara}[1]{\subsubsection{#1}}
\newcommand{\ourM}{{Res2Net}}
\newcommand{\sArt}{{state-of-the-art~}}
\newcommand{\tabSpace}{\vspace{6pt}}

\def\ie{\emph{i.e.,~}}
\def\eg{\emph{e.g.,~}}
\def\etc{\emph{etc}}
\def\etal{{\em et al.}}

\def\etal{{\em et al.}}
\newcommand{\tabFormat}{\centering \renewcommand{\arraystretch}{1.05}}
\usepackage[normalem]{ulem}

\RequirePackage{silence}
\graphicspath{{./figures/}{./CAM/}{./SAL/}{./SEG/}}

\begin{document}

\title{\ourM: A New Multi-scale Backbone Architecture}

\author{
Shang-Hua Gao$^{*}$,~
Ming-Ming Cheng$^{*}$,~
Kai Zhao,~
Xin-Yu Zhang,~
Ming-Hsuan Yang,~
and Philip Torr \\
\thanks{*Equal contribution}
\IEEEcompsocitemizethanks{
\IEEEcompsocthanksitem S.H. Gao, M.M. Cheng, K. Zhao, and X.Y Zhang are with
the TKLNDST, College of Computer Science, Nankai University,
Tianjin 300350, China.
\IEEEcompsocthanksitem M.H. Yang is with UC Merced.
\IEEEcompsocthanksitem P. Torr is with Oxford University.
\IEEEcompsocthanksitem M.M. Cheng is the corresponding author (cmm@nankai.edu.cn).
}
}
%
\markboth{IEEE Transactions on Pattern Analysis and Machine Intelligence, Vol. 43, No. 2, Feb.~2021}%
{Gao \MakeLowercase{\textit{et al.}}: 
\ourM: A New Multi-scale Backbone Architecture}

\IEEEtitleabstractindextext{%
\begin{abstract}
\justifying
Representing features at multiple scales is of great importance
for numerous vision tasks.
Recent advances in backbone convolutional neural networks (CNNs)
continually demonstrate stronger multi-scale representation ability,
leading to consistent performance gains on a wide range of applications.
However, most existing methods represent the multi-scale features
in a layer-wise manner.
In this paper, we propose a novel building block for CNNs,
namely \ourM,
by constructing hierarchical residual-like connections
within one single residual block.
The \ourM~represents multi-scale features at a granular level
and increases the range of receptive fields for each network layer.
The proposed \ourM~block can be plugged into the \sArt backbone CNN models,
\eg ResNet, ResNeXt, and DLA.
We evaluate the \ourM~block on all these models and demonstrate
consistent performance gains over baseline models on widely-used datasets,
\eg CIFAR-100 and ImageNet.
Further ablation studies and experimental results on representative computer
vision tasks, \ie object detection, class activation mapping,
and salient object detection,
further verify the superiority of the \ourM~over the \sArt baseline methods.
%
The source code and trained models are available on 
\url{https://mmcheng.net/res2net/}.

\end{abstract}
}

\maketitle
\IEEEdisplaynontitleabstractindextext
\IEEEpeerreviewmaketitle

\IEEEraisesectionheading{\section{Introduction}\label{sec:introduction}}
\IEEEPARstart{V}{isual} patterns occur at multi-scales in natural scenes
as shown in \figref{fig:receptive_field}.
First, objects may appear with different sizes in a single image,
\eg the sofa and cup are of different sizes.
Second, the essential contextual information of an object may occupy a much
larger area than the object itself.
For instance, we need to rely on the big table as context to better tell
whether the small black blob placed on it is a cup or a pen holder.
Third, perceiving information from different scales is essential for
understanding parts as well as objects for tasks 
such as fine-grained classification and semantic segmentation.
Thus, it is of critical importance to design good features for 
multi-scale stimuli for visual cognition tasks,
including image classification \cite{krizhevsky2012imagenet},
object detection \cite{ren2015faster},
attention prediction \cite{selvaraju2017grad},
target tracking \cite{zhang2017multi},
action recognition \cite{simonyan2014two},
semantic segmentation \cite{chen2018deeplab},
salient object detection \cite{hou2017deeply,BorjiCVM2019},
object proposal \cite{ren2015faster,BingObjCheng2018},
skeleton extraction \cite{zhao2018hifi}, 
stereo matching \cite{nie2019multi},
and edge detection \cite{xie2015holistically,liu2017richer}.

\CheckRmv{
\begin{figure}[t]
  \centering
  \includegraphics[width=\linewidth]{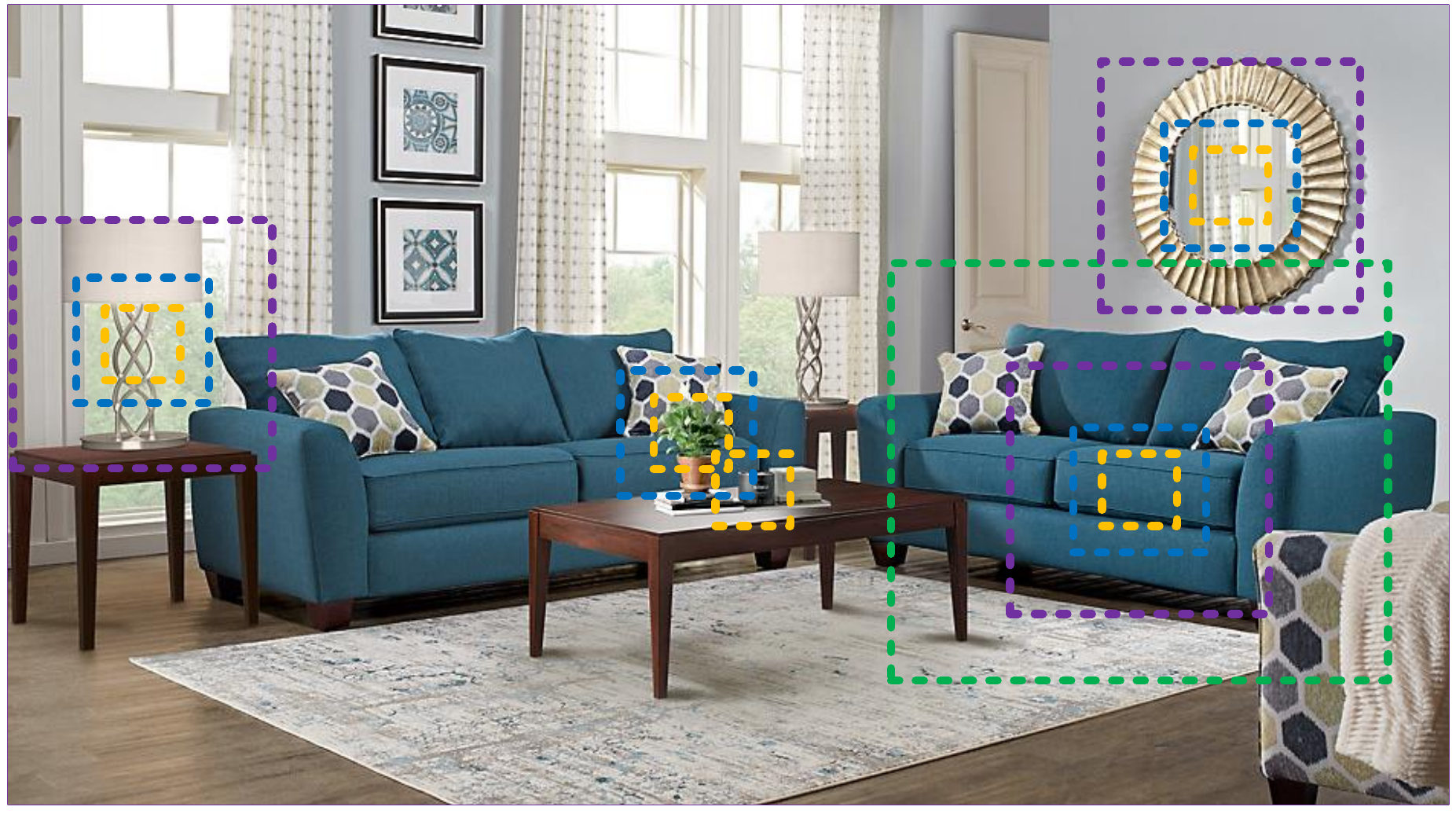}
  \caption{Multi-scale representations are essential for various vision tasks,
  	such as perceiving boundaries, regions,
  	and semantic categories of the target objects.
  	Even for the simplest recognition tasks,
  	perceiving information from very different scales is essential 
  	to understand parts, objects (\eg sofa, table, and cup in this example),
  	and their surrounding context
  	(\eg `on the table' context contributes to recognizing the black blob).
  }\label{fig:receptive_field}
\end{figure}
}

Unsurprisingly, multi-scale features have been widely used in 
both conventional feature design 
\cite{belongie2002shape,lowe2004distinctive}
and deep learning
\cite{szegedy2015going,chen2017dual}.
Obtaining multi-scale representations in vision tasks
requires feature extractors to use a large range of receptive fields
to describe objects/parts/context at different scales.
Convolutional neural networks (CNNs) naturally learn coarse-to-fine
multi-scale features through a stack of convolutional operators.
Such inherent multi-scale feature extraction ability of CNNs
leads to effective representations for solving numerous vision tasks.
How to design a more efficient network architecture is the key to further
improving the performance of CNNs.

In the past few years, several backbone networks, \eg
\cite{krizhevsky2012imagenet,simonyan2014very,szegedy2015going,he2016deep,
huang2017densely,Chollet_2017_CVPR,xie2017aggregated,chen2017dual,
yu2018deep,hu2018senet},
have made significant advances in numerous vision tasks with 
\sArt performance.
%
%
Earlier architectures such as AlexNet~\cite{krizhevsky2012imagenet} and
VGGNet~\cite{simonyan2014very} stack convolutional operators, 
making the data-driven learning of multi-scale features feasible.
The efficiency of multi-scale ability was subsequently improved by
using conv layers with different kernel size
(\eg InceptionNets~\cite{szegedy2015going,szegedy2016rethinking,
szegedy2017inception}),
residual modules (\eg ResNet~\cite{he2016deep}),
shortcut connections (\eg DenseNet~\cite{huang2017densely}),
and hierarchical layer aggregation (\eg DLA~\cite{yu2018deep}).
The advances in backbone CNN architectures have demonstrated
a trend towards more effective and efficient multi-scale representations.

\CheckRmv{
\begin{figure}[t]
  \centering
  \begin{overpic}[width=\linewidth]{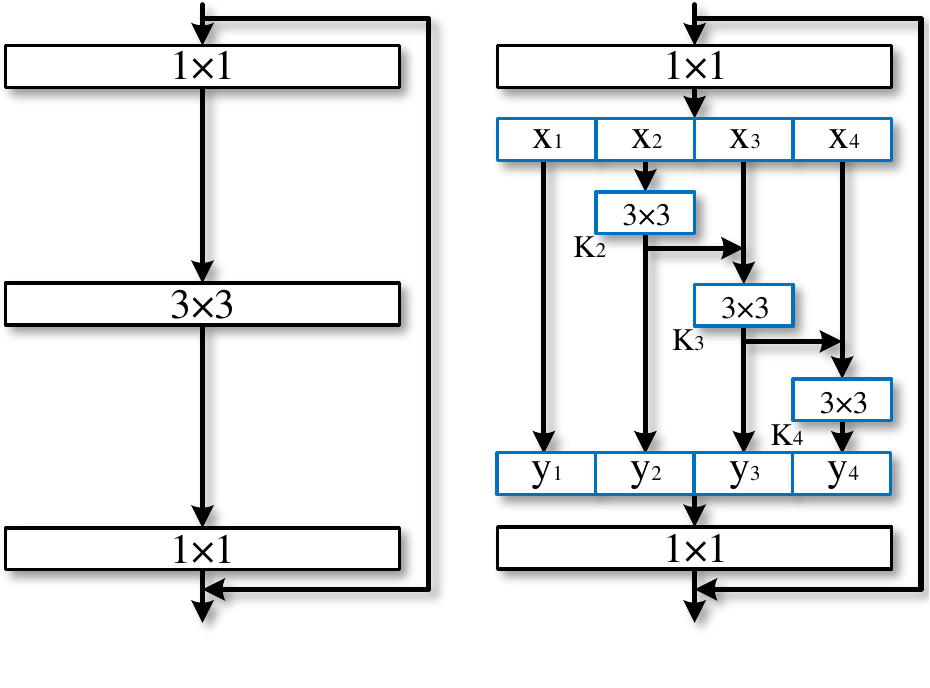}
     \put(7.5,2.5){(a) Bottleneck block}
     \put(57,2.5){(b) \ourM~module}
  \end{overpic}
  \caption{Comparison between the bottleneck block and the proposed
  	\ourM~module (the scale dimension $s=4$).
  }\label{fig:structure}
\end{figure}
}

In this work, we propose a simple yet efficient multi-scale processing approach.
Unlike most existing methods that enhance the
layer-wise multi-scale representation strength of CNNs,
we improve the multi-scale representation ability at a more granular level.
Different from some concurrent works~\cite{chen2019drop,chen2018biglittle,cheng2019high} 
that improve the multi-scale ability
by utilizing features with different resolutions,
 the multi-scale of our proposed method refers to 
the multiple available receptive fields at a more granular level.
To achieve this goal, we replace the $3 \times 3$ filters\footnote{
Convolutional operators and filters are used interchangeably.}
of $n$ channels, with a set of smaller filter groups, each with $w$ channels
(without loss of generality we use $n = s \times w$).
As shown in \figref{fig:structure},
these smaller filter groups are connected in a hierarchical residual-like style
to increase the number of scales that the output features can represent.
Specifically, we divide input feature maps into several groups.
A group of filters first extracts features from a group of input feature maps.
Output features of the previous group are then sent to
the next group of filters along with another group of input feature maps.
This process repeats several times until all input feature maps are processed.
Finally, feature maps from all groups are concatenated
and sent to another group of $1 \times 1$ filters to fuse information
altogether.
Along with any possible path in which input features are transformed to output features,
the equivalent receptive field increases whenever it passes 
a $3 \times 3$ filter,
resulting in many equivalent feature scales due to combination effects.

The \ourM~strategy exposes a \textbf{new dimension}, namely
\emph{scale} (the number of feature groups in the \ourM~block),
as an essential factor in addition to existing dimensions of
depth \cite{simonyan2014very}, width\footnote{
Width refers to the number of channels in a layer as in
\cite{Zagoruyko2016WRN}.},
and cardinality \cite{xie2017aggregated}.
We state in \secref{sec:compare_dimentions} that increasing
scale is more effective than increasing other dimensions.

Note that the proposed approach exploits the multi-scale
potential at a more granular level,
which is orthogonal to existing methods that utilize layer-wise operations.
Thus, the proposed building block, namely~\emph{\ourM~module},
can be easily plugged into many existing CNN architectures.
Extensive experimental results show that the \emph{\ourM~module} can further improve
the performance of state-of-the-art CNNs,
\eg ResNet~\cite{he2016deep}, ResNeXt~\cite{xie2017aggregated},
and DLA~\cite{yu2018deep}.

\section{Related Work}

\subsection{Backbone Networks}

Recent years have witnessed numerous backbone networks
\cite{krizhevsky2012imagenet,simonyan2014very,szegedy2015going,he2016deep,
huang2017densely,Chollet_2017_CVPR,xie2017aggregated,yu2018deep},
achieving \sArt performance in various vision tasks
with stronger multi-scale representations.
As designed, CNNs are equipped with basic multi-scale 
feature representation ability
since the input information follows a fine-to-coarse fashion.
%
The AlexNet~\cite{krizhevsky2012imagenet} stacks filters
sequentially and achieves significant performance gain
over traditional methods for visual recognition.
However, due to the limited network depth and kernel size of filters,
the AlexNet has only a relatively small receptive field.
The VGGNet~\cite{simonyan2014very} increases the network depth and uses filters
with smaller kernel size.
A deeper structure can expand the receptive fields,
which is useful for extracting features from a larger scale.
It is more efficient to enlarge the receptive field
by stacking more layers than using large kernels.
As such, the VGGNet provides a stronger multi-scale representation model
than AlexNet,
with fewer parameters.
However, both AlexNet and VGGNet stack filters directly, 
which means each feature layer has a relatively fixed receptive field.

Network in Network (NIN)~\cite{lin2013network} inserts multi-layer 
perceptrons as micro-networks into the large network 
to enhance model discriminability for local patches within the receptive field.
The 1 $\times$ 1 convolution introduced in NIN has been a popular module to fuse features.
The GoogLeNet~\cite{szegedy2015going}
utilizes parallel filters with different kernel sizes
to enhance the multi-scale representation capability.
However, such capability is often limited by the computational constraints
due to its limited parameter efficiency.
The Inception Nets~\cite{szegedy2016rethinking,szegedy2017inception}
stack more filters in each path of the parallel paths in the GoogLeNet
to further expand the receptive field.
On the other hand, the ResNet~\cite{he2016deep} introduces short connections
to neural networks,
thereby alleviating the gradient vanishing problem while
obtaining much deeper network structures.
During the feature extraction procedure,
short connections allow different combinations of convolutional operators,
resulting in a large number of equivalent feature scales.
Similarly, densely connected layers in the DenseNet~\cite{huang2017densely}
enable the network to process objects in a very wide range of scales.
DPN~\cite{chen2017dual} combines the ResNet with DenseNet to 
enable feature re-usage ability of ResNet and the feature exploration 
ability of DenseNet.
The recently proposed DLA~\cite{yu2018deep} method combines layers in a 
tree structure.
The hierarchical tree structure enables the network to obtain even stronger
layer-wise multi-scale representation capability.

\subsection{Multi-scale Representations for Vision Tasks}

Multi-scale feature representations of CNNs are of great importance to
a number of vision tasks including object detection~\cite{ren2015faster},
face analysis~\cite{bulat2017far,najibi2017ssh},
edge detection~\cite{liu2017richer},
semantic segmentation~\cite{chen2018deeplab},
salient object detection~\cite{Liu2019PoolSal,Zhao2019RgbdSal}, 
and skeleton detection~\cite{zhao2018hifi},
boosting the model performance of those fields.

\myPara{Object detection.}
Effective CNN models need to locate objects of different scales in a scene.
Earlier works such as the R-CNN~\cite{girshick2014rich} mainly rely on
the backbone network, \ie VGGNet~\cite{simonyan2014very},
to extract features of multiple scales.
He \etal~propose an SPP-Net approach~\cite{he2015spatial} that utilizes
spatial pyramid pooling after the backbone network
to enhance the multi-scale ability.
The Faster R-CNN method~\cite{ren2015faster} further proposes
the region proposal networks to generate bounding boxes with various scales.
Based on the Faster R-CNN, the FPN~\cite{lin2017feature} approach 
introduces feature pyramid
to extract features with different scales from a single image.
The SSD method~\cite{liu2016ssd} utilizes feature maps from different stages
to process visual information at different scales.

\myPara{Semantic segmentation.}

Extracting essential contextual information of objects requires CNN models
to process features at various scales for effective semantic segmentation.
Long \etal~\cite{long2015fully} propose one of the earliest methods that
enables multi-scale representations of the fully convolutional network (FCN)
for semantic segmentation task.
In DeepLab,
Chen \etal~\cite{chen2018deeplab,chen2017rethinking}
introduces cascaded atrous convolutional module to
expand the receptive field further while preserving spatial resolutions.
More recently, global context information is aggregated from
region-based features via the pyramid pooling scheme in the
PSPNet~\cite{Zhao2017PSP}.

\myPara{Salient object detection.}

Precisely locating the salient object regions in an image
requires an understanding of both large-scale context information
for the determination of object saliency, 
and small-scale features to localize object boundaries accurately~\cite{zhao2019optimizing}.
Early approaches~\cite{borji2015salient} utilize handcrafted
representations of global contrast~\cite{cheng2015global} or
multi-scale region features \cite{WangDRFI2017}.
Li \etal \cite{li2015visual} propose one of the earliest methods that
enables multi-scale deep features for salient object detection.
Later, multi-context deep learning \cite{zhao2015saliency}
and multi-level convolutional features \cite{zhang2017amulet}
are proposed for improving salient object detection.
More recently, Hou \etal \cite{hou2017deeply} introduce dense
short connections among stages to provide rich multi-scale
feature maps at each layer for salient object detection.

\subsection{Concurrent Works} \label{sec:Concurrent_works}

Recently, there are some concurrent works aiming at improving the performance
by utilizing the multi-scale features
~\cite{chen2019drop,chen2018biglittle,cheng2019high,SunZJCXLMWLW19}.
Big-Little Net~\cite{chen2018biglittle} is a multi-branch network 
composed of branches with different computational complexity.
Octave Conv~\cite{chen2019drop} decomposes the standard convolution into 
two resolutions to process features at different frequencies.
MSNet~\cite{cheng2019high} utilizes a high-resolution network to learn
high-frequency residuals by using the up-sampled low-resolution features 
learned by a low-resolution network.
Other than the low-resolution representations in current works, 
the HRNet~\cite{SunXLW19,SunZJCXLMWLW19} introduces high-resolution
representations in the network and repeatedly performs multi-scale 
fusions to strengthen high-resolution representations.
One common operation in
~\cite{chen2019drop,chen2018biglittle,cheng2019high,SunXLW19,SunZJCXLMWLW19} 
is that they all use pooling or up-sample
to re-size the feature map to $2^n$ times of the original scale 
to save the computational budget while maintaining or even improving performance.
While in the Res2Net block, the hierarchical
residual-like connections within a single residual
block module enable the variation of receptive fields 
at a more granular level to capture details and global features.
Experimental results show that Res2Net module can be integrated 
with those novel network designs to further boost the performance.




\section{\ourM}

\subsection{\ourM~Module}

The bottleneck structure shown in \figref{fig:structure}(a)
is a basic building block in many modern backbone CNNs architectures,
\eg ResNet~\cite{he2016deep}, ResNeXt~\cite{xie2017aggregated},
and DLA~\cite{yu2018deep}.
Instead of extracting features using a group of $3 \times 3$ filters
as in the bottleneck block,
we seek alternative architectures with stronger multi-scale feature
extraction ability,
while maintaining a similar computational load.
Specifically, we replace a group of $3\times 3$ filters
with smaller groups of filters,
while connecting different filter groups in a hierarchical residual-like style.
Since our proposed neural network module involves 
\textbf{res}idual-like connections
within a single \textbf{res}idual block, we name it \emph{\ourM}.


\figref{fig:structure} shows the differences between the bottleneck block 
and the proposed \ourM~module.
After the $1 \times 1$ convolution,
we evenly split the feature maps into $s$ feature map subsets,
denoted by $\mathbf{x}_i$, where $i \in \{1, 2, ..., s\}$.
Each feature subset $\mathbf{x}_i$ has the same spatial size
but $1/s$ number of channels compared with the input feature map.
Except for $\mathbf{x}_1$, each $\mathbf{x}_i$ has a corresponding
$3 \times 3$ convolution,
denoted by $\mathbf{K}_i()$.
We denote by $\mathbf{y}_i$ the output of $\mathbf{K}_i()$.
The feature subset $\mathbf{x}_i$ is added with the output of
$\mathbf{K}_{i-1}()$, and then fed into $\mathbf{K}_i()$.
To reduce parameters while increasing $s$,
we omit the $3 \times 3$ convolution for $\mathbf{x}_1$.
Thus, $\mathbf{y}_i$ can be written as:
\CheckRmv{
\begin{eqnarray}
  \mathbf{y}_i=
  \left\{
   \begin{array}{lll}
     \mathbf{x}_i  & i=1;\\
     \mathbf{K}_i(\mathbf{x}_i) & i=2;\\
     \mathbf{K}_i(\mathbf{x}_i+ \mathbf{y}_{i-1}) & 2<i \leqslant s.
   \end{array}
   \right.
   \label{eq:out_i}
\end{eqnarray}
}

Notice that each $3 \times 3$ convolutional operator $\mathbf{K}_i()$
could potentially receive feature information from all feature splits
$\{\mathbf{x}_j, j \leq i\}$. 
Each time a feature split $\mathbf{x}_j$ goes through a $3 \times 3$
convolutional operator,
the output result can have a larger receptive field than $\mathbf{x}_j$.
Due to the combinatorial explosion effect,
the output of the \ourM~module contains a different number and 
different combination of receptive field sizes/scales.

In the \ourM~module,
splits are processed in a multi-scale fashion,
which is conducive to the extraction of both global and local information.
To better fuse information at different scales,
we concatenate all splits and pass them through a $1 \times 1$ convolution.
The split and concatenation strategy
can enforce convolutions
to process features more effectively.
To reduce the  number of parameters, we omit the convolution for the first split,
which can also be regarded as a form of feature reuse.

In this work,  we use $s$ as a control parameter of the \emph{scale} dimension.
Larger $s$ potentially allows features 
with richer receptive field sizes to be learnt,
with negligible computational/memory overheads introduced by concatenation.

\subsection{Integration with Modern Modules}

\CheckRmv{
\begin{figure}[t]
  \begin{overpic}[width=\linewidth]{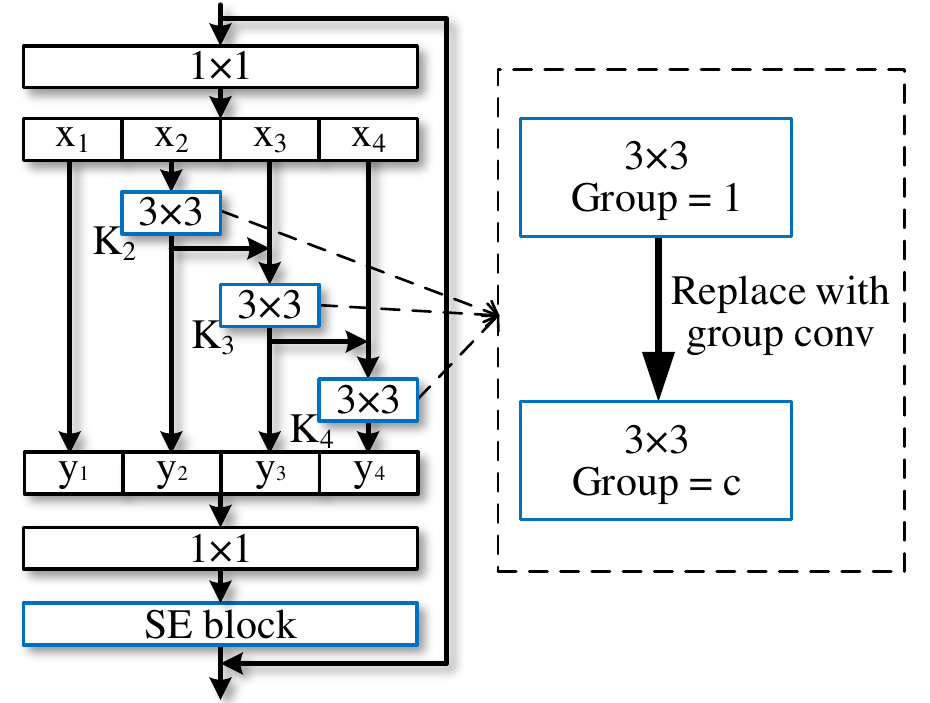}
  \end{overpic}
  \caption{The \ourM~module can be integrated with
    the dimension cardinality~\cite{xie2017aggregated} (replace conv with group conv)
    and SE~\cite{hu2018senet} blocks.
  }\label{fig:improved_structure}
\end{figure}
}

Numerous neural network modules have been proposed in recent years,
including cardinality dimension introduced by Xie \etal \cite{xie2017aggregated},
as well as squeeze and excitation (SE) block presented by
Hu \etal \cite{hu2018senet}.
The proposed \ourM~module introduces the scale dimension
that is orthogonal to  these improvements.
As shown in \figref{fig:improved_structure},
we can easily integrate the cardinality dimension ~\cite{xie2017aggregated}
and the SE block~\cite{hu2018senet} with the proposed \ourM~module.

\myPara{Dimension cardinality.}
The dimension cardinality indicates the number of groups within a
filter~\cite{xie2017aggregated}.
This dimension changes filters from single-branch to multi-branch and
improves the representation ability of a CNN model.
In our design, we can replace the 3 $\times$ 3 convolution
with the 3 $\times$ 3 group convolution,
where $c$ indicates the number of groups.
Experimental comparisons between the scale dimension and cardinality
are presented in \secref{sec:Experiments_ImageNet} and
\secref{sec:compare_dimentions}.

\myPara{SE block.}
A SE block adaptively re-calibrates channel-wise feature responses
by explicitly modelling inter-dependencies among channels~\cite{hu2018senet}.
Similar to \cite{hu2018senet},
we add the SE block right before the residual connections of the \ourM~module.
Our \ourM~module can benefit from the integration of the SE block,
which we have experimentally demonstrated in
\secref{sec:Experiments_ImageNet} and \secref{sec:Experiments_CIFAR}.

\subsection{Integrated Models}

Since the proposed \ourM~module does not have specific requirements of
the overall network structure
and the multi-scale representation ability of the \ourM~module is orthogonal to
the layer-wise feature aggregation models of CNNs,
we can easily integrate the proposed \ourM~module into the \sArt~models,
such as ResNet~\cite{he2016deep},
ResNeXt~\cite{xie2017aggregated}, DLA~\cite{yu2018deep} 
and Big-Little Net~\cite{chen2018biglittle}.
The corresponding models are referred to as
\ourM, Res2NeXt, Res2Net-DLA, and bLRes2Net-50,  respectively.

The proposed scale dimension is orthogonal to the cardinality
\cite{xie2017aggregated}
dimension and width \cmm{\cite{he2016deep}} dimension of prior work.
Thus, after the \emph{scale} is set,
we adjust the value of cardinality and width to maintain the overall
model complexity similar to its counterparts.
We do not focus on reducing the model size in this work
since it requires more meticulous designs such as depth-wise separable
convolution~\cite{Ma_2018_ECCV},
model pruning~\cite{han2015learning}, 
and model compression~\cite{cheng2017survey}.

For experiments on the ImageNet~\cite{russakovsky2015imagenet} dataset,
we mainly use the ResNet-50~\cite{he2016deep}, ResNeXt-50~\cite{xie2017aggregated},
DLA-60~\cite{yu2018deep}, and bLResNet-50 \cite{chen2018biglittle}
as our baseline models.
The complexity of the proposed model is approximately equal to 
those of the baseline models,
whose number of parameters is around $25M$ and the number of FLOPs
for an image of $224 \times 224$ pixels is around $4.2G$
for 50-layer networks.
For experiments on the CIFAR~\cite{krizhevsky2009learning} dataset,
we use the ResNeXt-29, 8c$\times$64w~\cite{xie2017aggregated} 
as our baseline model.
Empirical evaluations and discussions of the proposed models with respect to
model complexity are presented in \secref{sec:compare_dimentions}.

\section{Experiments}\label{sec:experiments}

\subsection{Implementation Details}
We implement the proposed models using the Pytorch framework.
For fair comparisons, we use the Pytorch implementation of ResNet~\cite{he2016deep},
ResNeXt~\cite{xie2017aggregated}, DLA~\cite{yu2018deep} as well as 
bLResNet-50~\cite{chen2018biglittle},
and only replace the original bottleneck block with the proposed \ourM~module.
Similar to prior work, on the ImageNet dataset~\cite{russakovsky2015imagenet},
each image is of 224$\times$224 pixels randomly cropped from
a re-sized image.
We use the same data argumentation strategy as \cite{he2016deep,szegedy2016rethinking}.
Similar to~\cite{he2016deep},
we train the network using SGD with weight decay 0.0001, momentum 0.9,
and a mini-batch of 256 on 4 Titan Xp GPUs.
The learning rate is initially set to 0.1 and divided by 10 every 30 epochs.

All models for the ImageNet, including the baseline and  proposed models,
are trained for 100 epochs with the same training and data argumentation strategy.
For testing, we use the same image cropping method as~\cite{he2016deep}.
On the CIFAR dataset, we use the implementation of
ResNeXt-29~\cite{xie2017aggregated}.
For all tasks, we use the original implementations of baselines and
only replace the backbone model with the proposed \ourM.

\newcommand{\ResNet}{{ResNet-50~\cite{he2016deep}}}
\newcommand{\ResNeXt}{{ResNeXt-50~\cite{xie2017aggregated}}}
\newcommand{\DLA}{{DLA-60~\cite{yu2018deep}}}
\newcommand{\DLAX}{{DLA-X-60~\cite{yu2018deep}}}
\newcommand{\ResNetSE}{{SENet-50~\cite{hu2018senet}}}
\newcommand{\ResNeXtSE}{{SENeXt-50~\cite{hu2018senet}}}
\newcommand{\InceptionV}{{InceptionV3~\cite{szegedy2016rethinking}}}
\newcommand{\BLResNet}{{bLResNet-50~\cite{chen2018biglittle}}}

\CheckRmv{
\begin{table}[tbp]
  \tabFormat
  \setlength{\tabcolsep}{2.3mm}
  \caption{Top-1 and Top-5 test error on the ImageNet dataset.}
  \tabSpace
  \begin{tabular}{lcc}\toprule
                    &top-1 err. ($\%$)&top-5 err. ($\%$)\\ \midrule
    \ResNet         & 23.85          & 7.13             \\
    \ourM-50        & \textbf{22.01} & \textbf{6.15}    \\ \midrule
    \InceptionV     & 22.55          & 6.44             \\
    \ourM-50-299    & \textbf{21.41} & \textbf{5.88}    \\ \midrule 
    \ResNeXt        & 22.61          & 6.50             \\
    Res2NeXt-50     & \textbf{21.76} & \textbf{6.09}    \\ \midrule
    \DLA            & 23.32          & 6.60             \\
    \ourM-DLA-60    & \textbf{21.53} & \textbf{5.80}    \\
    \DLAX           & 22.19          & 6.13             \\
    Res2NeXt-DLA-60 & \textbf{21.55} & \textbf{5.86}    \\ \midrule
    \ResNetSE       & 23.24          & 6.69             \\
    SE-\ourM-50     & \textbf{21.56} & \textbf{5.94}    \\ \midrule
    \BLResNet       & 22.41          &   -              \\
    bLRes2Net-50    & \textbf{21.68} & 6.00             \\ \midrule
    Res2Net-v1b-50  & 19.73          & 4.96             \\ 
    Res2Net-v1b-101 & 18.77          & 4.64             \\ \midrule
    Res2Net-200-SSLD~\cite{ma2019paddlepaddle}& \textbf{14.87}  &  -       \\ 
    \bottomrule
  \end{tabular}
  \label{tab:imagenet_prec}
\end{table}
}

\subsection{ImageNet}\label{sec:Experiments_ImageNet}

We conduct experiments on the ImageNet dataset~\cite{russakovsky2015imagenet},
which contains 1.28 million training images and 50k validation images  
from 1000 classes.
We construct the models with  approximate 50 layers for performance evaluation
against the \sArt methods.
More ablation studies are conducted on the CIFAR dataset.

\myPara{Performance gain.}
\tabref{tab:imagenet_prec} shows the top-1 and top-5 test error 
on the ImageNet dataset.
For simplicity, all \ourM~models in \tabref{tab:imagenet_prec} 
have the scale $s=4$.
The \ourM-50 has an improvement of 1.84$\%$ on top-1 error over the ResNet-50.
The Res2NeXt-50 achieves a 0.85$\%$ improvement in terms of top-1 error 
over the ResNeXt-50.
Also, the Res2Net-DLA-60 outperforms the DLA-60 by
1.27$\%$ in terms of top-1 error.
The Res2NeXt-DLA-60 outperforms the DLA-X-60 by
0.64$\%$ in terms of top-1 error.
The SE-Res2Net-50 has an improvement of 1.68$\%$ over the SENet-50.
bLRes2Net-50 has an improvement of 0.73$\%$ in terms of top-1 error 
over the bLResNet-50.
The \ourM~module further enhances the multi-scale ability of bLResNet 
at a granular level even bLResNet is designed to utilize features 
with different scales as discussed in~\secref{sec:Concurrent_works}.
Note that the ResNet \cite{he2016deep}, ResNeXt \cite{xie2017aggregated},
SE-Net \cite{hu2018senet}, bLResNet~\cite{chen2018biglittle}, 
and DLA \cite{yu2018deep} are the state-of-the-art CNN models.
Compared with these strong baselines,
models integrated with the \ourM~module still have consistent performance gains.

We also compare our method against the 
InceptionV3~\cite{szegedy2016rethinking} model,
which  utilizes parallel filters with different kernel combinations.
For fair comparisons, we use the~\ResNet~as the baseline model and
train our model with the input image size of 299$\times$299 pixels,
as used in the InceptionV3 model.
The proposed \ourM-50-299 outperforms InceptionV3 by 1.14$\%$ on top-1 error.
We conclude that the hierarchical residual-like connection of the \ourM~module is more
effective than the parallel filters of InceptionV3 when processing multi-scale information.
While the combination pattern of filters in InceptionV3 is dedicatedly designed,
the \ourM~module presents a simple but effective combination pattern.




\myPara{Going deeper with \ourM.}

Deeper networks have been shown to have stronger representation
capability~\cite{he2016deep,xie2017aggregated} for vision tasks.
To validate our model with greater depth,
we compare the classification performance of the \ourM~and the ResNet,
both with 101 layers.
As shown in~\tabref{tab:imagenet_prec_101},
the \ourM-101 achieves significant performance gains over
the ResNet-101 with 1.82$\%$ in terms of top-1 error.
Note that the \ourM-50 has the performance gain of 1.84$\%$ in terms
of top-1 error over the ResNet-50.
These results show that the proposed module with additional dimension scale
can be integrated with deeper models to achieve better performance.
We also compare our method with the DenseNet~\cite{huang2017densely}.
Compared with the DenseNet-161, the best performing model of the
officially provided DenseNet family,
the \ourM-101 has an improvement of 1.54$\%$ in terms of top-1 error.

\newcommand{\DenseNet}{{DenseNet-161~\cite{huang2017densely}}}
\newcommand{\ResNetl}{{ResNet-101~\cite{he2016deep}}}

\CheckRmv{
\begin{table}[tbp]
  \tabFormat
  \setlength{\tabcolsep}{2.2mm}
  \caption{Top-1 and Top-5 test error ($\%$) of deeper networks on the 
    ImageNet dataset.
  }\tabSpace
  \begin{tabular}{lcc}\toprule
              &  top-1 err.    & top-5 err.    \\ \midrule
    \DenseNet & 22.35          & 6.20           \\
    \ResNetl  & 22.63          & 6.44           \\
    \ourM-101 & \textbf{20.81} & \textbf{5.57}  \\
  \bottomrule
  \end{tabular}
  \label{tab:imagenet_prec_101}
\end{table}
}

\myPara{Effectiveness of scale dimension.}

\newcommand{\RowsCapt}[1]{{\multirow{3}{*}{\begin{tabular}[l]{@{}l@{}} \ourM-50\\ #1 \\ complexity) \end{tabular}}}}

\CheckRmv{
\begin{table}[tbp]
  \tabFormat
  \setlength{\tabcolsep}{1.4mm}
  \caption{Top-1 and Top-5 test error ($\%$) of Res2Net-50 with
     different scales on the ImageNet dataset.
     Parameter $w$ is the width of filters,
     and $s$ is the number of scale, as described in \eqref{eq:out_i}.
  }
  \tabSpace
  \begin{tabular}{lccccc}\toprule
               & Setting &FLOPs  &Runtime & top-1 err. & top-5 err.\\ \midrule
    ResNet-50  & 64w           & 4.2G   & 149ms  & 23.85 & 7.13 \\ \midrule
    \RowsCapt{(
    Preserved} & 48w$\times$2s & 4.2G & 148ms  & 22.68 & 6.47 \\
               & 26w$\times$4s & 4.2G & 153ms  & 22.01 & 6.15 \\
               & 14w$\times$8s & 4.2G & 172ms  & 21.86 & 6.14 \\ \midrule
    \RowsCapt{(
    Increased} & 26w$\times$4s & 4.2G  & -     & 22.01 & 6.15 \\
               & 26w$\times$6s & 6.3G  & -     & 21.42 & 5.87 \\
               & 26w$\times$8s & 8.3G  & -     & 20.80 & 5.63 \\ \midrule
    \ourM-50-L & 18w$\times$4s & 2.9G  & 106ms & 22.92 & 6.67 \\\bottomrule
  \end{tabular}
\label{tab:imagenet_vary_scale}
\end{table}
}

To validate our proposed dimension scale,
we experimentally analyze the effect of different scales.
As shown in ~\tabref{tab:imagenet_vary_scale},
the performance increases with the increase of scale.
With the increase of scale, the \ourM-50 with 14w$\times$8s
achieves performance gains over the ResNet-50 with 1.99$\%$ in terms of top-1 error.
Note that with the preserved complexity,
the width of $\mathbf{K}_i()$ decreases with the increase of scale.
We further evaluate the performance gain of increasing scale with increased model complexity.
The \ourM-50 with 26w$\times$8s
achieves significant performance gains over the ResNet-50 with 3.05$\%$ in terms of top-1 error.
A \ourM-50 with 18w$\times$4s also outperforms the ResNet-50 
by 0.93$\%$ in terms of top-1 error with only 69$\%$ FLOPs.
\tabref{tab:imagenet_vary_scale} shows the Runtime under different scales, 
which is the average time to infer the ImageNet validation set with the 
size of 224 $\times$ 224.
Although the feature splits $\{\mathbf{y}_i\}$ need to be computed sequentially 
due to hierarchical connections,
the extra run-time introduced by Res2Net module can often be ignored.
Since the number of available tensors in a GPU is limited,
there are typically sufficient parallel computations within a single GPU
clock period for the typical setting of Res2Net, \ie $s=4$.

\myPara{Stronger representation with ResNet.}
To further explore the multi-scale representation ability of Res2Net,
we follow the ResNet v1d~\cite{he2019bag} to modify Res2Net,
and train the model with data augmentation techniques \ie CutMix~\cite{yun2019cutmix}.
The modified version of Res2Net, namely Res2Net~v1b,
greatly improve the classification performance on ImageNet as shown in~\tabref{tab:imagenet_prec}.
Res2Net v1b further improve the model performance on downstream tasks. 
We show the performance of Res2Net~v1b
on object detection, instance segmentation, key-points estimation in~\tabref{tab:object_detection}, 
~\tabref{tab:instance_segmantation}, and~\tabref{tab:keypoint}, respectively.

The stronger multi-scale representation of Res2Net 
has been verified on many downstream tasks \ie vectorized road extraction~\cite{VecRoad_20CVPR},
object detection~\cite{li2020generalized},
weakly supervised semantic segmentation~\cite{21PAMI_InsImgDatasetWSIS}, salient object detection~\cite{GaoEccv20Sal100K},
interactive image segmentation~\cite{fClick20CVPR}, video recognition~\cite{li2020tea}, 
concealed object detection~\cite{fan2021concealed},
and medical segmentation~\cite{wu2020jcs,fan2020pranet,fan2020inf}.
Semi-supervised knowledge distillation solution~\cite{ma2019paddlepaddle} 
can also be applied to Res2Net,
to achieve the 85.13$\%$ top.1 acc. on ImageNet.

\subsection{CIFAR}
\label{sec:Experiments_CIFAR}

We also conduct some experiments on the CIFAR-100
dataset~\cite{krizhevsky2009learning},
which contains 50k training images and 10k testing images from 100 classes.
The \emph{ResNeXt-29, 8c$\times$64w}~\cite{xie2017aggregated} is used
as the baseline model.
We only replace the original basic block with our proposed \ourM~module
while keeping other configurations unchanged.
\tabref{tab:cifar_prec} shows the top-1 test error and model size 
on the CIFAR-100 dataset.
Experimental results show that our method surpasses the baseline
and other methods with fewer parameters.
Our proposed ~\emph{Res2NeXt-29, 6c$\times$24w$\times$6s} outperforms
the baseline by 1.11$\%$.
\emph{Res2NeXt-29, 6c$\times$24w$\times$4s} even outperforms
the \emph{ResNeXt-29, 16c$\times$64w} with only 35$\%$ parameters.
We also achieve better performance with fewer parameters,
compared with DenseNet-BC ($k = 40$).
Compared with \emph{Res2NeXt-29, 6c$\times$24w$\times$4s},
\emph{Res2NeXt-29, 8c$\times$25w$\times$4s} achieves a better result
with more width and cardinality,
indicating that the dimension scale is orthogonal to dimension
width and cardinality.
We also integrate the recently proposed SE block into our structure.
With fewer parameters,
our method still outperforms the \emph{ResNeXt-29, 8c$\times$64w-SE} baseline.

\newcommand{\WideResNet}{{Wide ResNet~\cite{Zagoruyko2016WRN}}}
\newcommand{\ResNeXtBase}{{ResNeXt-29, 8c$\times$64w~\cite{xie2017aggregated} (base)}}
\newcommand{\ResNeXtLarge}{{ResNeXt-29, 16c$\times$64w~\cite{xie2017aggregated}}}
\newcommand{\ResNeXtSEs}{{ResNeXt-29, 8c$\times$64w-SE~\cite{hu2018senet}}}
\newcommand{\DenseNetBC}{{DenseNet-BC (k = 40)~\cite{huang2017densely}}}
\newcommand{\RessNetTms}[3]{Res2NeXt-29, {#1}c$\times${#2}w$\times${#3}s}
\newcommand{\RessNetTmsSE}[3]{Res2NeXt-29, {#1}c$\times${#2}w$\times${#3}s-SE}

\CheckRmv{
\begin{table}[tbp]
  \tabFormat
  \setlength{\tabcolsep}{1.5mm}
  \caption{Top-1 test error ($\%$) and model size on the CIFAR-100 dataset.
      Parameter $c$ indicates the value of cardinality, and $w$ is the width of filters.}
  \tabSpace
  \begin{tabular}{lcc} \toprule
                        & Params & top-1 err.     \\ \midrule
  \WideResNet           & 36.5M  & 20.50          \\
  \ResNeXtBase          & 34.4M  & 17.90          \\
  \ResNeXtLarge         & 68.1M  & 17.31          \\
  \DenseNetBC           & 25.6M  & 17.18          \\
  \RessNetTms{6}{24}{4} & 24.3M  & 16.98          \\ 
  \RessNetTms{8}{25}{4} & 33.8M  & 16.93          \\ 
  \RessNetTms{6}{24}{6} & 36.7M  & \textbf{16.79} \\ 
  \midrule
  \ResNeXtSEs           & 35.1M  & 16.77          \\
  \RessNetTmsSE{6}{24}{4} & 26.0M  & 16.68        \\ 
  \RessNetTmsSE{8}{25}{4} & 34.0M  & 16.64        \\ 
  \RessNetTmsSE{6}{24}{6} & 36.9M & \textbf{16.56}\\ \bottomrule 
  \end{tabular}
  \label{tab:cifar_prec}
\end{table}
}

\newcommand{\addFig}[1]{{\includegraphics[height=.135\textwidth]{#1.pdf}}}

\CheckRmv{
\begin{figure*}[t]
  \centering
  \small
  \renewcommand{\arraystretch}{0.5}
  \setlength{\tabcolsep}{0.2mm}
  \begin{tabular}{cccccccc}
   \rotatebox[origin=l]{90}{~~~~~ResNet-50}&
   \addFig{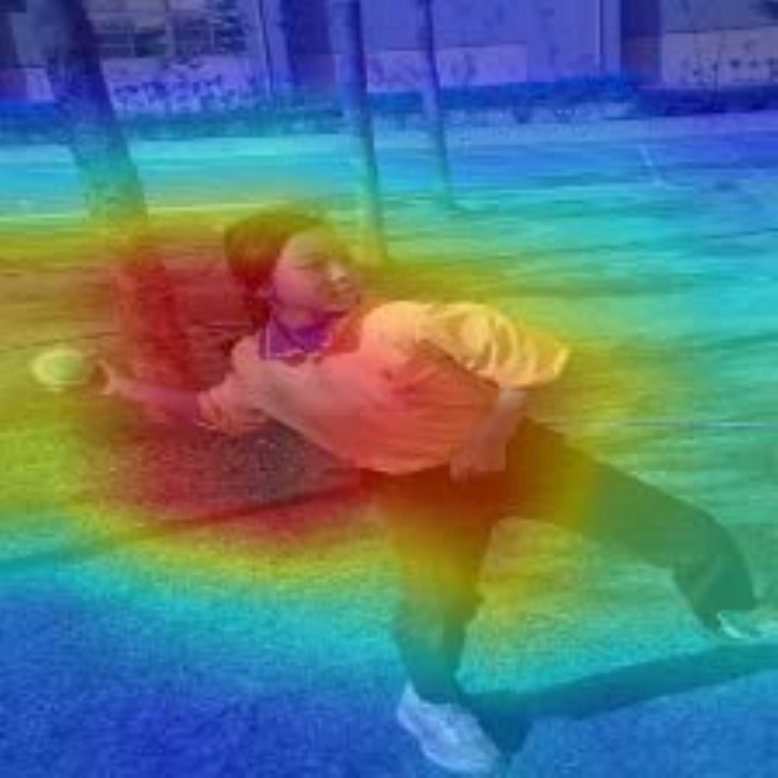}&
    \addFig{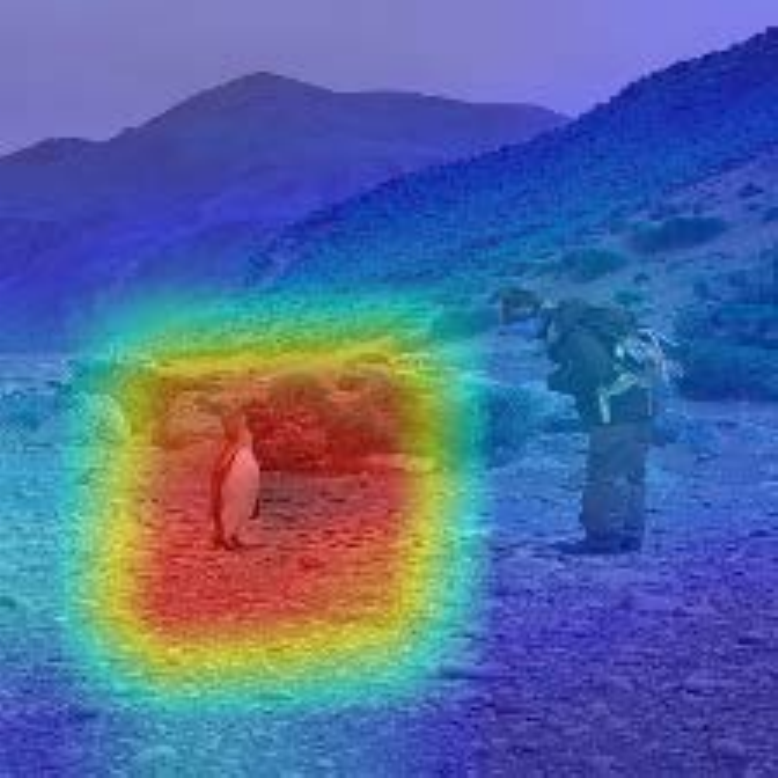}&
    \addFig{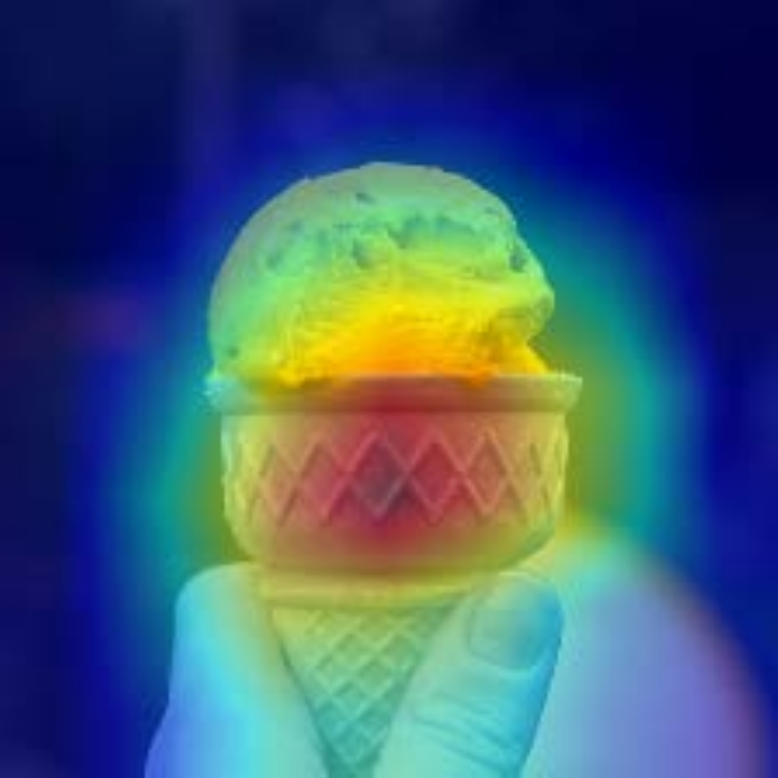}&
   \addFig{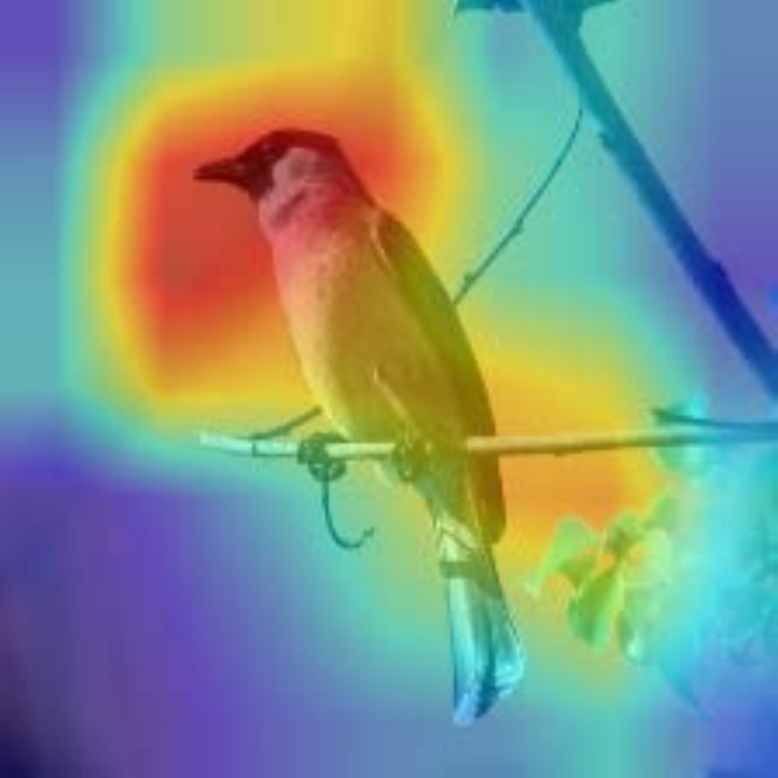}&
    \addFig{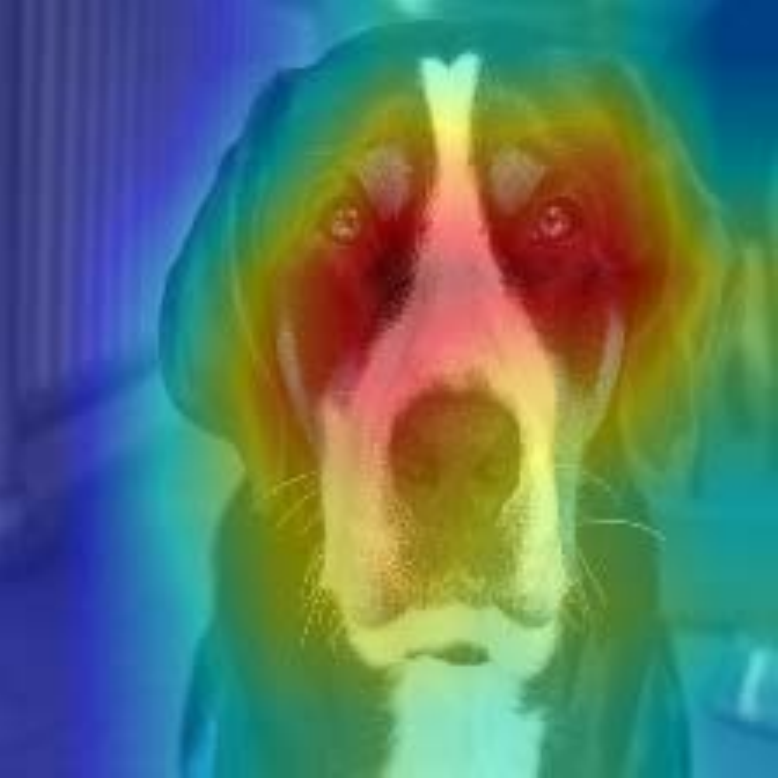}&
   \addFig{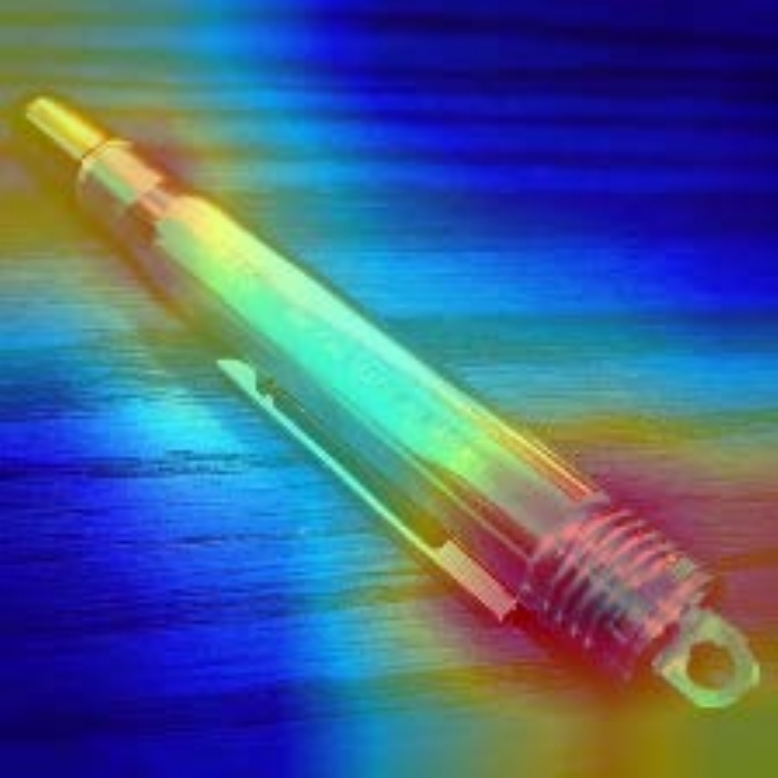}&
   \addFig{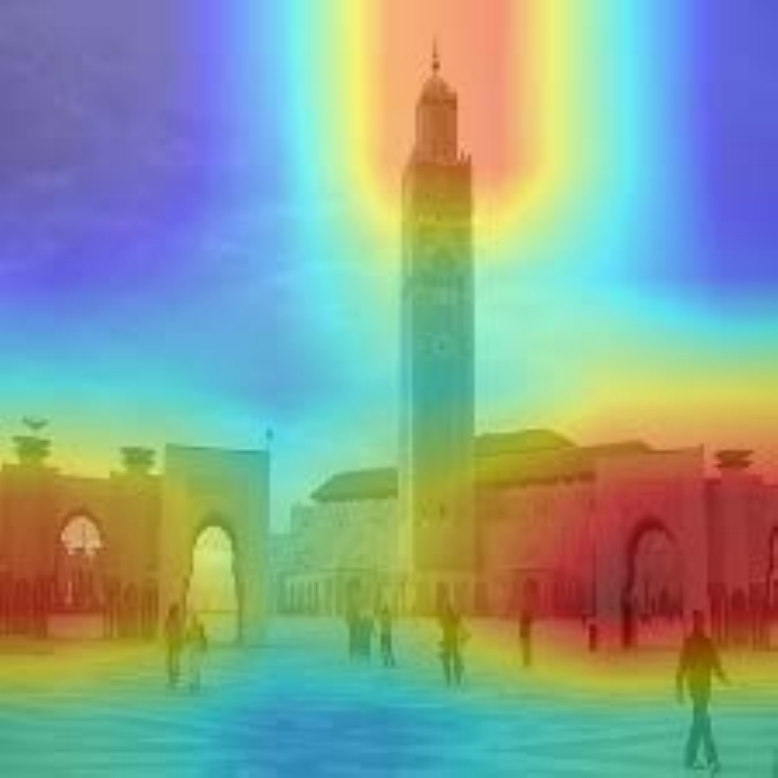}
   \\
   \rotatebox[origin=l]{90}{~~~~~Res2Net-50}&
   \addFig{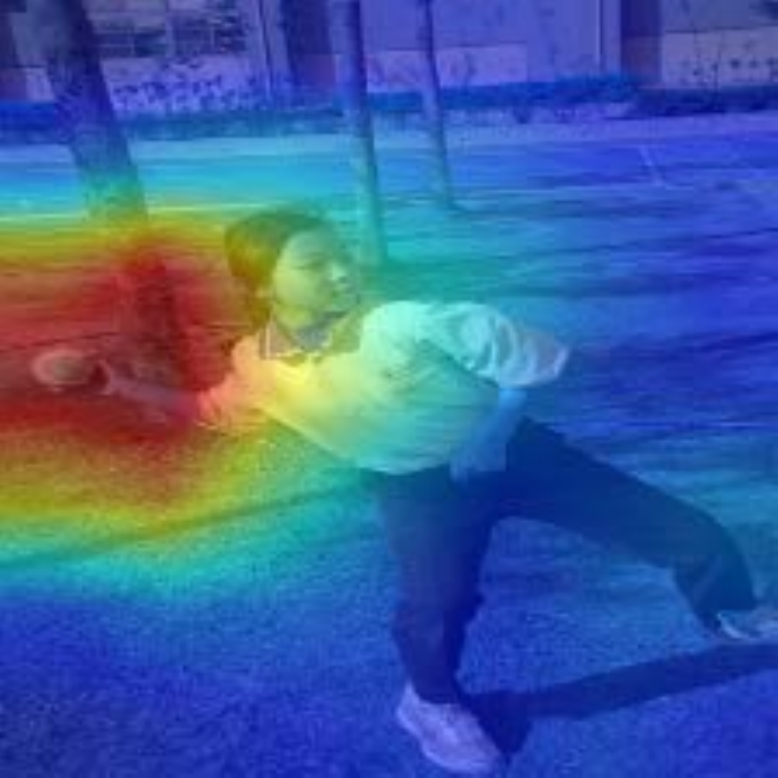}&
   \addFig{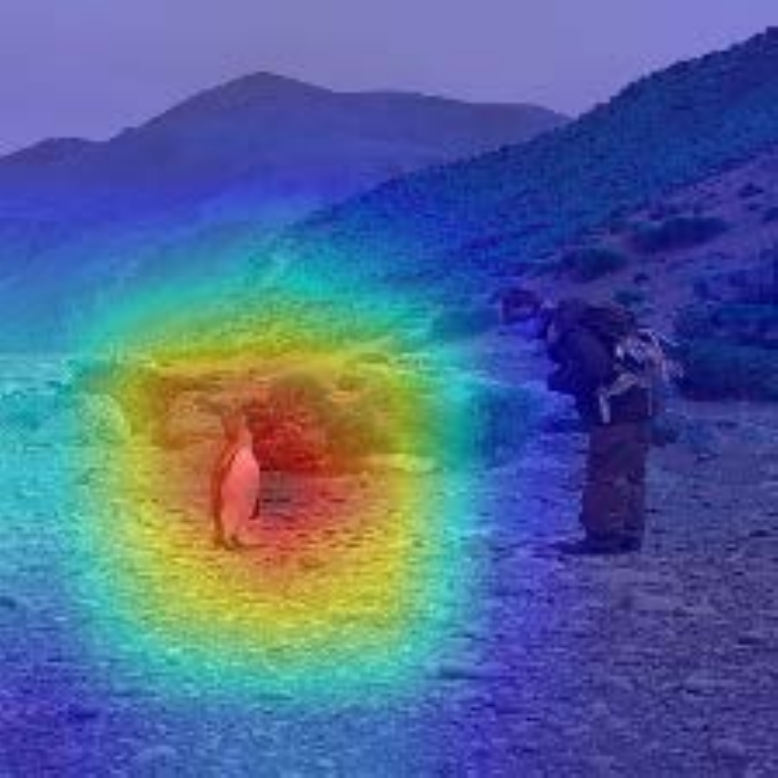}&
  \addFig{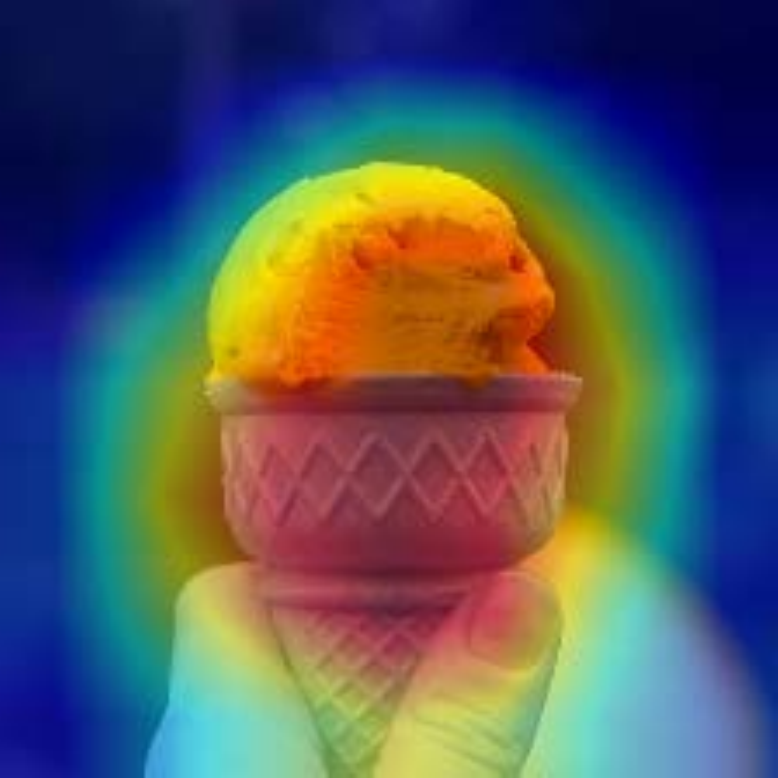}&
   \addFig{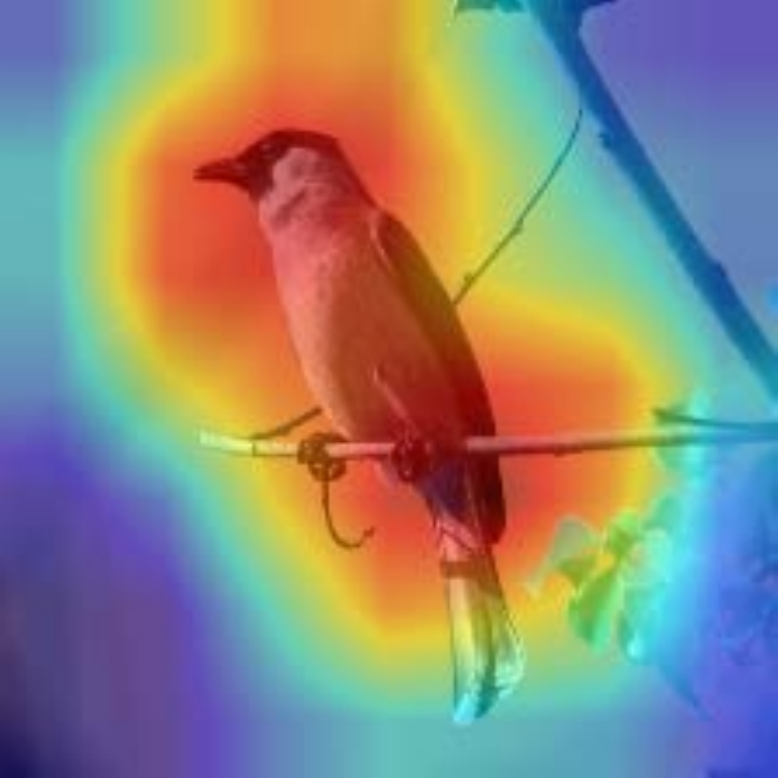}&
   \addFig{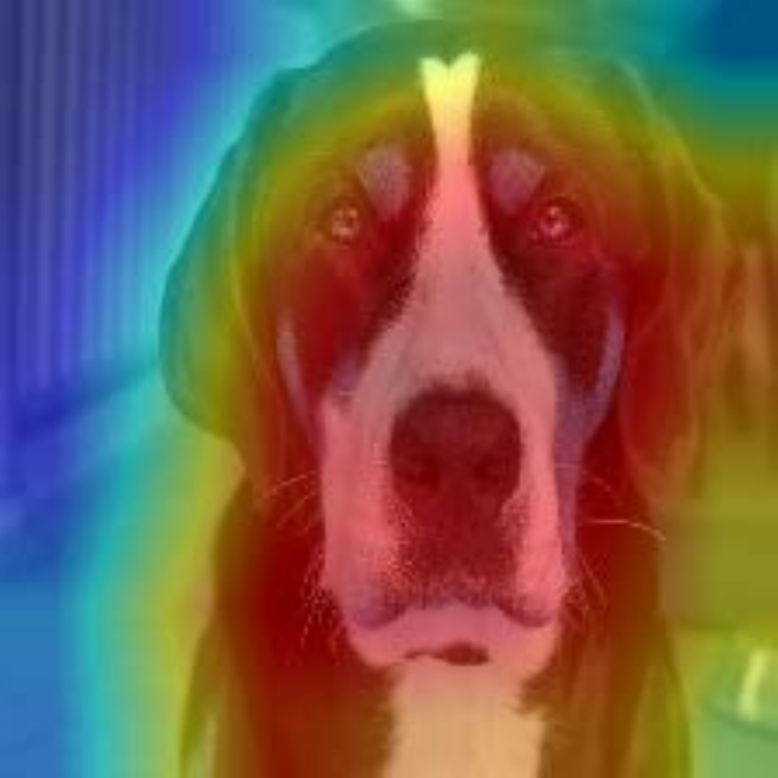} &
   \addFig{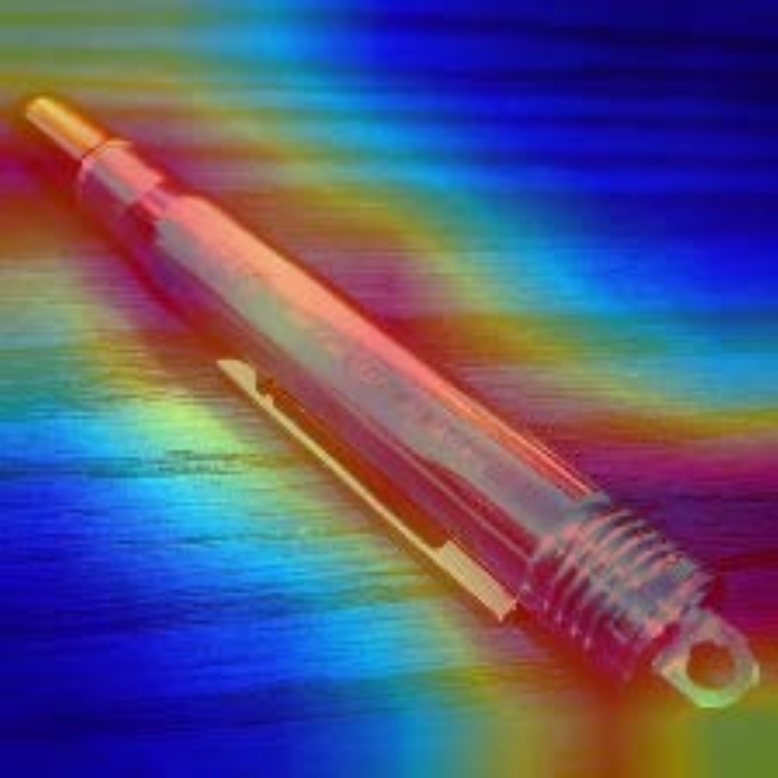}&
   \addFig{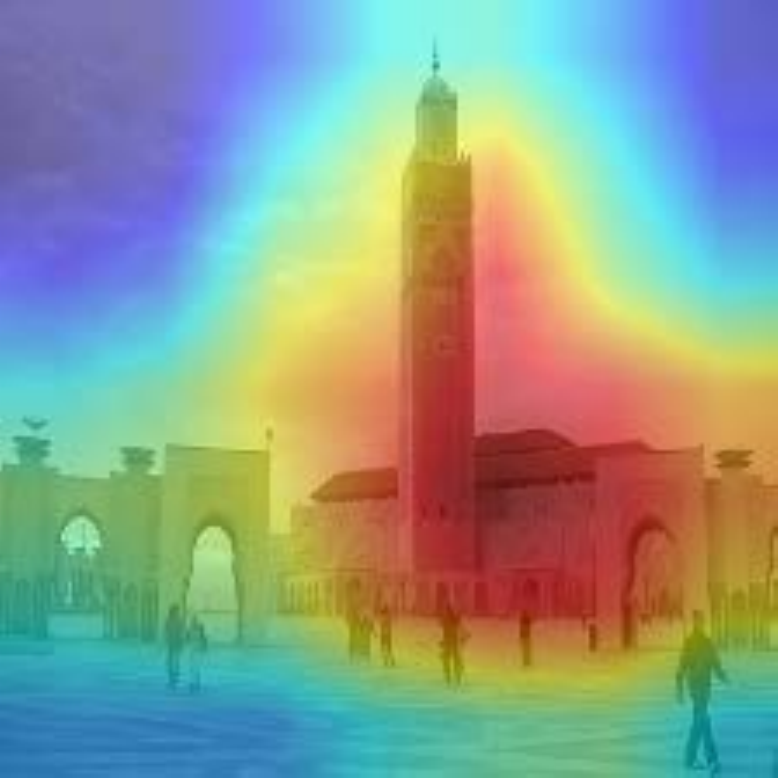}
   \\
   &Baseball & Penguin & Ice cream & Bulbul & Mountain dog & Ballpoint & Mosque \\
  \end{tabular}
  \caption{Visualization of class activation mapping \cite{selvaraju2017grad},
  	using ResNet-50 and Res2Net-50 as backbone networks.}
  \label{fig:cam}
\end{figure*}
}

\subsection{Scale Variation}
\label{sec:compare_dimentions}

Similar to Xie \etal \cite{xie2017aggregated},
we evaluate the test performance of the baseline model by increasing
different CNN dimensions,
including scale (\eqref{eq:out_i}), cardinality \cite{xie2017aggregated},
and depth \cite{simonyan2014very}.
While increasing model capacity using one dimension,
we fix all other dimensions.
A series of networks are trained and evaluated under these changes.
Since \cite{xie2017aggregated} has already shown that increasing cardinality
is more effective than increasing width,
we only compare the proposed dimension scale with cardinality and depth.

\figref{fig:cmp_scale_card} shows the test precision on the CIFAR-100 dataset
with regard to the model size.
The depth, cardinality, and scale of the baseline model are
$29, 6$ and $1$, respectively.
Experimental results suggest that scale is an effective dimension
to improve model performance,
which is consistent with what we have observed on the ImageNet dataset
in \secref{sec:Experiments_ImageNet}.
Moreover, increasing scale is more effective than other dimensions,
resulting in quicker performance gains.
As described in ~\eqref{eq:out_i} and \figref{fig:structure},
for the case of scale $s=2$,
we only increase the model capacity by adding more parameters
of $1 \times 1$ filters.
Thus, the model performance of $s=2$ is slightly worse
than that of increasing cardinality.
For $s = 3,4$, the combination effects of our hierarchical residual-like
structure produce a rich set of equivalent scales,
resulting in significant performance gains.
However, the models with scale 5 and 6 have limited performance gains,
about which we assume that the image in the CIFAR dataset is too small (32$\times$32)
to have many scales.

\CheckRmv{
\begin{figure}[t]
  \centering
  \begin{overpic}[width=\linewidth]{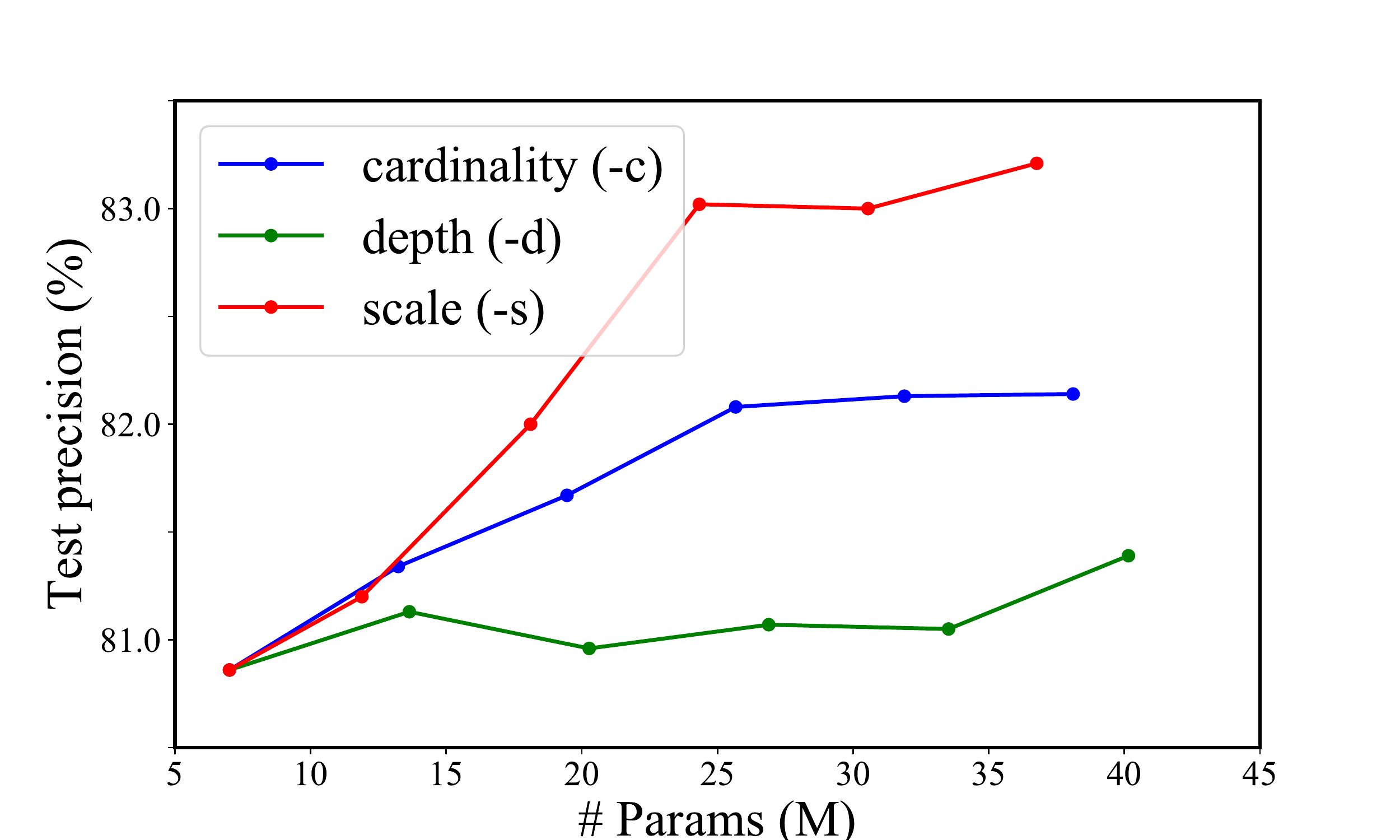}
    \put(22,21.5){2$s$} \put(39.5,31){3$s$} \put(54,48){4$s$}
    \put(66,47){5$s$} \put(80,51){6$s$}
    \put(30,20){12$c$} \put(43,25){18$c$} \put(56,31.5){24$c$}
    \put(70,32.5){30$c$} \put(84,33){36$c$}
    \put(29,14.5){56$d$} \put(44,12){83$d$} \put(58,13.5){110$d$}
    \put(73,13.5){137$d$} \put(87,19.5){164$d$}
    \put(14,10){29$d$-6$c$-1$s$} 
  \end{overpic}
  \caption{Test precision on the CIFAR-100 dataset with regard to the model size,
  	by changing cardinality (ResNeXt-29), depth (ResNeXt),
  	and scale (Res2Net-29).
  }\label{fig:cmp_scale_card}
\end{figure}
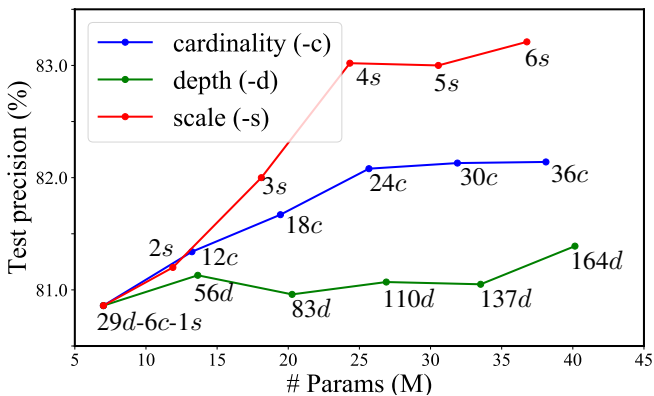
}

\subsection{Class Activation Mapping}

To understand the multi-scale ability of the \ourM, 
we visualize the class activation mapping (CAM) using 
Grad-CAM~\cite{selvaraju2017grad},
which is commonly used to localize the discriminative regions for 
image classification.
In the visualization examples shown in \figref{fig:cam},
stronger CAM areas are covered with lighter colors.
Compared with ResNet, the \ourM~based CAM results have more concentrated 
activation maps on small objects such as `baseball' and `penguin'.
Both methods have similar activation maps on the middle size objects,
such as `ice cream'.
Due to stronger multi-scale ability, 
the \ourM~has activation maps that tend to cover the whole object on big objects 
such as `bulbul', `mountain dog', `ballpoint', and `mosque', 
while activation maps of ResNet only cover parts of objects.
Such ability of precisely localizing CAM region makes the \ourM~ 
potentially valuable for object region mining in 
weakly supervised semantic segmentation tasks \cite{AdversErasingCVPR2017}.

\newcommand{\Rows}[1]{\multirow{2}{*}{#1}}
\newcommand{\threeRows}[1]{\multirow{4}{*}{#1}}
\CheckRmv{
\begin{table}[tbp]
  \tabFormat
  \setlength{\tabcolsep}{3.6mm}
  \caption{Object detection results on the PASCAL VOC07 and COCO
    datasets,
    measured using AP ($\%$) and AP@IoU=0.5 ($\%$).
    The \ourM~has similar complexity compared with its counterparts.
  }\tabSpace
  \begin{tabular}{llcc} \toprule
    Dataset     & Backbone  &  AP  & AP@IoU=0.5             \\ \midrule
    \Rows{VOC07}& ResNet-50 & 72.1 & -                      \\
                & \ourM-50  & \textbf{74.4} & -             \\ \midrule
    \threeRows{COCO} & ResNet-50 & 31.1          & 51.4      \\
                & \ourM-50  & 33.7      & 53.6 \\ 
                & \ourM-v1b-101 & \textbf{43.0} & \textbf{63.5} \\ \bottomrule
  \end{tabular}
\label{tab:object_detection}
\end{table}
}

\newcommand{\RowsT}[1]{{\multirow{3}{*}{\begin{tabular}[c]{@{}c@{}}#1\\ ($\%$)\end{tabular}}}}
\CheckRmv{
\begin{table}[tbp]
  \tabFormat
  \setlength{\tabcolsep}{2.1mm}
  \caption{Average Precision (AP) and Average Recall (AR)
    of object detection with different sizes on the COCO dataset.
  }\tabSpace
  \begin{tabular}{lccccc} \toprule
              &            &\multicolumn{4}{c}{Object size }\\ \cline{3-6}
              &            & Small & Medium & Large & All  \\  \midrule
    ResNet-50 & \RowsT{AP} & 13.5 & 35.4 & 46.2 & 31.1 \\
    \ourM-50  &            & 14.0 & 38.3 & 51.1 & 33.7 \\
    Improve.  &            & +0.5 & +2.9 & +4.9 & +2.6 \\ \midrule
    ResNet-50 & \RowsT{AR} & 21.8 & 48.6 & 61.6 & 42.8 \\
    \ourM-50  &            & 23.2 & 51.1 & 65.3 & 45.0 \\
    Improve.  &            & +1.4 & +2.5 & +3.7 & +2.2 \\ \bottomrule
  \end{tabular}
  \label{tab:object_detection_size}
\end{table}
}

\subsection{Object Detection}
\label{sec:object_det}
For object detection task,
we validate the \ourM~on the PASCAL VOC07 \cite{everingham2010pascal}
and MS COCO \cite{lin2014microsoft} datasets,
using Faster R-CNN~\cite{ren2015faster} as the baseline method.
We use the backbone network of ResNet-50 vs. \ourM-50,
and follow all other implementation details of \cite{ren2015faster} for a fair comparison.
%
%
\tabref{tab:object_detection} shows the object detection results.
On the PASCAL VOC07 dataset,
the \ourM-50 based model outperforms its counterparts by 
$2.3\%$ on average precision (AP).
On the COCO dataset, the \ourM-50 based model outperforms its counterparts by
$2.6\%$ on AP, and $2.2\%$ on AP@IoU=0.5.

We further test the AP and average recall (AR) scores for objects
of different sizes as shown in \tabref{tab:object_detection_size}.
Objects are divided into three categories based on the size,
according to~\cite{lin2014microsoft}.
The \ourM~based model has a large margin of improvement over its counterparts by
$0.5\%$, $2.9\%$, and $4.9\%$
on AP for small, medium, and large objects, respectively.
The improvement of AR for small, medium, and large objects are
$1.4\%$, $2.5\%$, and $3.7\%$, respectively.
%
%
Due to the strong multi-scale ability,
the \ourM~based models can cover a large range of receptive fields,
boosting the performance on objects of different sizes.

\subsection{Semantic Segmentation}\label{sec:semantic_seg}
\CheckRmv{
\begin{table}[tbp]
  \tabFormat
  \setlength{\tabcolsep}{2.3mm}
  \centering
  \tabSpace
  \caption{Performance of semantic segmentation on PASCAL 
    VOC12 val set using Res2Net-50 with different scales.
    The Res2Net has similar complexity compared with its counterparts.
  }
  \begin{tabular}{lccccccc}\toprule
   Backbone        & Setting       & Mean IoU ($\%$)  \\ \midrule
   ResNet-50       & 64w           & 77.7             \\ \midrule
   \multirow{4}{*}
   {Res2Net-50}    & 48w$\times$2s & 78.2             \\
                   & 26w$\times$4s & \textbf{79.2}    \\
                   & 18w$\times$6s & 79.1             \\
                   & 14w$\times$8s & 79.0             \\ \midrule
   ResNet-101      & 64w           & 79.0             \\ 
   Res2Net-101     & 26w$\times$4s & \textbf{80.2}    \\ 
  \bottomrule
  \end{tabular}
  \label{tab:semantic_segmentation}
\end{table}
}

\newcommand{\addSegFig}[1]{{\includegraphics[height=.120\textwidth]{#1}}}
\newcommand{\addSegFigGT}[1]{\addSegFig{gt/#1.pdf}}
\newcommand{\addSegFigRR}[1]{\addSegFig{res2net/#1.pdf}}
\newcommand{\addSegFigR}[1]{\addSegFig{resnet/#1.pdf}}
\CheckRmv{
\begin{figure*}[t]
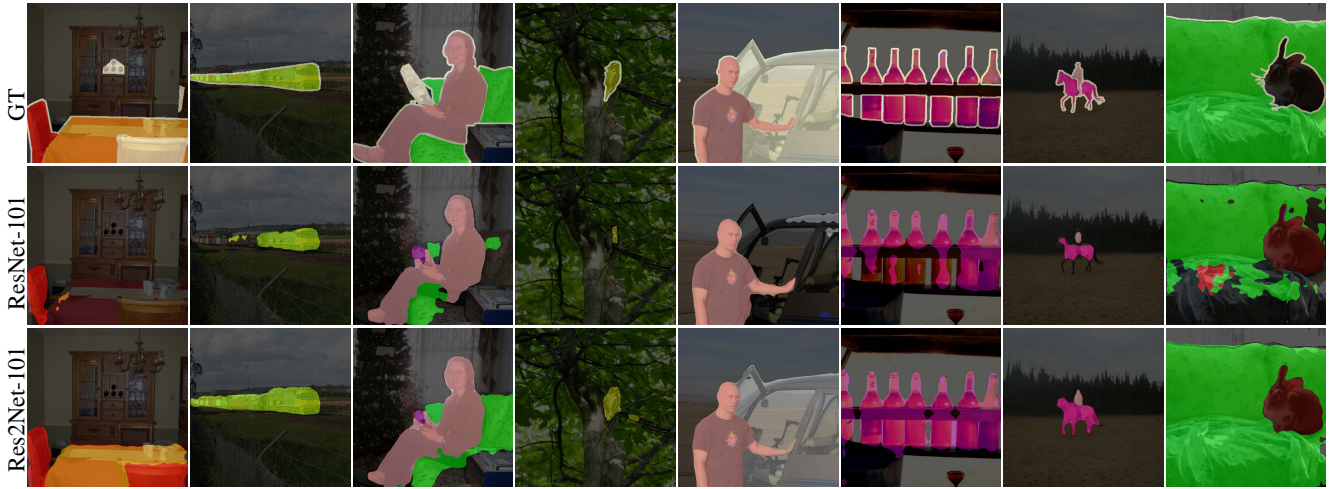

  \centering
  \small
  \renewcommand{\arraystretch}{0.5}
  \setlength{\tabcolsep}{0.2mm}
  \begin{tabular}{ccccccccc}
    \rotatebox[origin=l]{90}{~~~~~GT}&
    \addSegFigGT{54}&
    \addSegFigGT{636}&
    \addSegFigGT{307}&
    \addSegFigGT{1071}&
    \addSegFigGT{482}&
    \addSegFigGT{936}&
    \addSegFigGT{1422}&
    \addSegFigGT{1443}
    \\
   \rotatebox[origin=l]{90}{~ResNet-101}&
   \addSegFigR{54}&
   \addSegFigR{636}&
   \addSegFigR{307}&
   \addSegFigR{1071}&
   \addSegFigR{482}&
   \addSegFigR{936}&
   \addSegFigR{1422}&
   \addSegFigR{1443}
   \\
   \rotatebox[origin=l]{90}{~Res2Net-101}&
   \addSegFigRR{54}&
   \addSegFigRR{636}&
   \addSegFigRR{307}&
   \addSegFigRR{1071}&
   \addSegFigRR{482}&
   \addSegFigRR{936}&
   \addSegFigRR{1422}&
   \addSegFigRR{1443}
   \\
  \end{tabular}
  \caption{Visualization of semantic segmentation results~\cite{Chen_2018_ECCV},
  	using ResNet-101 and Res2Net-101 as backbone networks.}
  \label{fig:segvis}
\end{figure*}
}

Semantic segmentation requires a strong multi-scale ability of CNNs
to extract essential contextual information of objects.
We thus evaluate the multi-scale ability of \ourM~on the semantic segmentation task
using PASCAL VOC12 dataset~\cite{everingham2015pascal}.
We follow the previous work to use the augmented PASCAL VOC12
dataset~\cite{hariharan2011semantic}
which contains 10582 training images and 1449 val images.
We use the Deeplab v3+~\cite{Chen_2018_ECCV} as our segmentation method.
All implementations remain the same with Deeplab v3+~\cite{Chen_2018_ECCV}
except that the backbone network is replaced with ResNet and our proposed \ourM.
The output strides used in training and evaluation are both 16.
As shown in ~\tabref{tab:semantic_segmentation},
\ourM-50 based method outperforms its counterpart by 1.5$\%$ on mean IoU.
And \ourM-101 based method outperforms its counterpart by 1.2$\%$ on mean IoU.
Visual comparisons of semantic segmentation results on challenging examples 
are illustrated in \figref{fig:segvis}.
The \ourM~based method tends to segment all parts of objects 
regardless of object size.

\subsection{Instance Segmentation}

Instance segmentation is the combination of object detection 
and semantic segmentation.
It requires not only the correct detection of objects with various sizes in an image
but also the precise segmentation of each object.
As mentioned in \secref{sec:object_det} and \secref{sec:semantic_seg},
both object detection and semantic segmentation require a strong 
multi-scale ability of CNNs.
Thus, the multi-scale representation is quite beneficial to instance segmentation.
We use the Mask R-CNN~\cite{he2017mask} as the instance segmentation method, 
and replace the backbone network of ResNet-50 with our proposed~\ourM-50.
The performance of instance segmentation on MS COCO \cite{lin2014microsoft} dataset 
is shown in \tabref{tab:instance_segmantation}.
The \ourM-26w$\times$4s~based method outperforms its counterparts 
by 1.7$\%$ on $AP$ and 2.4$\%$ on $AP_{50}$.
The performance gains on objects with different sizes are also demonstrated.
The improvement of $AP$ for small, medium, and large objects are 
0.9$\%$, 1.9$\%$, and 2.8$\%$, respectively.
\tabref{tab:instance_segmantation} also shows the performance comparisons of \ourM~ 
under the same complexity with different scales.
The performance shows an overall upward trend  with the increase of scale.
Note that compared with the Res2Net-50-48w$\times$2s, the Res2Net-50-26w$\times$4s
has an improvement of 2.8 $\%$ on $AP_{L}$, while the Res2Net-50-48w$\times$2s 
has the same $AP_{L}$ compared with ResNet-50.
We assume that the performance gain on large objects is benefited from the extra scales.
When the scale is relatively larger, the performance gain is not obvious.
The Res2Net module is capable of learning a suitable range of receptive fields.
The performance gain is limited when the scale of objects in the image
is already covered by the available receptive fields in the Res2Net module.
With fixed complexity, the increased scale results in fewer channels for each receptive field, 
which may reduce the ability to process features of a particular scale.

\CheckRmv{
\begin{table}[tbp]
  \tabFormat
  \setlength{\tabcolsep}{0.9mm}
  \centering
  \caption{Performance of instance segmentation on the COCO dataset 
     using Res2Net-50 with different scales.
     The Res2Net has similar complexity compared with its counterparts.
  }\tabSpace
  \begin{tabular}{lccccccc}\toprule
     Backbone     & Setting       & $AP$ & $AP_{50}$ & $AP_{75}$ & $AP_{S}$ & $AP_{M}$ & $AP_{L}$ \\ \midrule
     ResNet-50    & 64w           & 33.9 & 55.2 & 36.0 & 14.8 & 36.0 & 50.9 \\ \midrule
     \multirow{4}{*}
     {Res2Net-50} & 48w$\times$2s & 34.2 & 55.6 & 36.3 & 14.9 & 36.8 & 50.9 \\
                  & 26w$\times$4s & 35.6 & \textbf{57.6} & 37.6 & \textbf{15.7} & 37.9 & \textbf{53.7} \\
                  & 18w$\times$6s & \textbf{35.7} & 57.5 & \textbf{38.1} & 15.4 & \textbf{38.1} & 53.7 \\
                  & 14w$\times$8s & 35.3 & 57.0 & 37.5 & 15.6 & 37.5 & 53.4 \\ \midrule
     \multirow{1}{*}
{Res2Net-v1b-101} & 64w           & 38.7 & 61.0 & 41.4 & 20.6 & 42.0 & 53.2  \\
      \bottomrule
  \end{tabular}
  \label{tab:instance_segmantation}
\end{table}
}

\subsection{Salient Object Detection}

\CheckRmv{
\begin{table}[t]
  \tabFormat
  \setlength{\tabcolsep}{1.9mm}
  \caption{Salient object detection results on different datasets,
   measured using F-measure and Mean Absolute Error (MAE).
   The \ourM~has similar complexity compared with its counterparts.
  }
  \tabSpace
  \begin{tabular}{llcc} \toprule
  Dataset        & Backbone   & F-measure$\uparrow$& MAE $\downarrow$ \\ \midrule
  \multirow{2}
  {*}{ECSSD}     & ResNet-50  & 0.910          & 0.065          \\
                 & Res2Net-50 & \textbf{0.926} & \textbf{0.056} \\ \midrule
  \multirow{2}
  {*}{PASCAL-S}  & ResNet-50  & 0.823          & 0.105          \\
                 & Res2Net-50 & \textbf{0.841} & \textbf{0.099} \\ \midrule
  \multirow{2}
  {*}{HKU-IS}    & ResNet-50  & 0.894          & 0.058          \\
                 & Res2Net-50 & \textbf{0.905} & \textbf{0.050} \\ \midrule
  \multirow{2}
  {*}{DUT-OMRON} & ResNet-50  & 0.748           & 0.092         \\
                 & Res2Net-50 & \textbf{0.800}  & \textbf{0.071}\\ \bottomrule
  \end{tabular}
  \label{tab:sod_result}
\end{table}
}

Pixel level tasks such as salient object detection also require
the strong multi-scale ability of CNNs to locate both the holistic objects
as well as their region details.
Here we use the latest method DSS~\cite{hou2017deeply} as our baseline.
For a fair comparison, we only replace the backbone with ResNet-50 and
our proposed \ourM-50, while keeping other configurations unchanged.
Following \cite{hou2017deeply}, we train those two models using the
MSRA-B dataset~\cite{liu2011learning},
and evaluate results on ECSSD~\cite{yan2013hierarchical}, PASCAL-S~\cite{li2014secrets},
HKU-IS~\cite{li2015visual}, and DUT-OMRON~\cite{yang2013saliency} datasets.
The F-measure and Mean Absolute Error (MAE) are used for evaluation.
As shown in~\tabref{tab:sod_result},
the \ourM~based model has a consistent improvement compared with its counterparts
on all datasets.
On the DUT-OMRON dataset (containing 5168 images), the \ourM~based model has a
$5.2\%$ improvement on F-measure and a $2.1\%$ improvement on MAE,
compared with ResNet based model.
The \ourM~based approach achieves greatest performance gain on the DUT-OMRON dataset,
since this dataset contains the most significant object size variation
compared with the other three datasets.
Some visual comparisons of salient object detection results on challenging
examples are illustrated in \figref{fig:salvis}.

\renewcommand{\addFig}[1]{{\includegraphics[width=.116\textwidth]{#1.pdf}}}
\newcommand{\addFigs}[1]{\addFig{img/#1}&\addFig{gt/#1}&\addFig{resnet/#1}&\addFig{res2net/#1}}
\CheckRmv{
\begin{figure}[t]
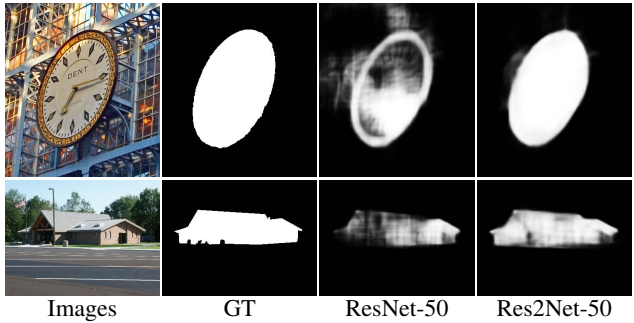

  \centering
  \small
  \setlength{\tabcolsep}{0.2mm}
  \renewcommand{\arraystretch}{0.5}
  \begin{tabular}{@{}cccc@{}}
    \addFigs{0424}\\
    \addFigs{sun_ahyszoiuweslyfqs}\\
    Images & GT & ResNet-50 & Res2Net-50 \\
  \end{tabular}
  \caption{Examples of salient object detection \cite{hou2017deeply} results,
  	using ResNet-50 and Res2Net-50 as backbone networks, respectively.}
  \label{fig:salvis}
\end{figure}
}

\subsection{Key-points Estimation}

Human parts are of different sizes, which requires the key-points estimation method
to locate human key-points with different scales.
To verify whether the multi-scale representation ability of Res2Net can benefit the task
of key-points estimation, we use the SimpleBaseline~\cite{Xiao_2018_ECCV} as
the key-points estimation method and only replace the backbone with the proposed Res2Net.
All implementations including the training and testing strategies 
remain the same with the SimpleBaseline~\cite{Xiao_2018_ECCV}.
We train the model using the COCO key-point detection dataset~\cite{lin2014microsoft},
and evaluate the model using the COCO validation set.
Following common settings, we use the same person detectors 
in SimpleBaseline~\cite{Xiao_2018_ECCV} for evaluation.
~\tabref{tab:keypoint} shows the performance of key-points estimation on 
the COCO validation set using Res2Net.
The Res2Net-50 and Res2Net-101 based models outperform baselines on $AP$ 
by 3.3$\%$ and 3.0$\%$, respectively.
Also, Res2Net based models have considerable performance gains on human with different scales
compared with baselines.
\CheckRmv{
\begin{table}[tbp]
    \tabFormat
    \setlength{\tabcolsep}{1.3mm}
    \centering
    \caption{Performance of key-points estimation on the COCO validation set.
             The Res2Net has similar complexity compared with its counterparts.}
    \tabSpace
    \begin{tabular}{lcccccc}\toprule
     Backbone   & $AP$ &$AP_{50}$&$AP_{75}$& $AP_{M}$& $AP_{L}$ \\ \midrule  
     ResNet-50  & 70.4 & 88.6    & 78.3    & 67.1    & 77.2     \\            
     Res2Net-50 & 71.5 & 89.0	 & 79.3	   & 68.2	 & 78.4     \\    
     ResNet-101 & 71.4 & 89.3    & 79.3    & 68.1    & 78.1     \\            
     Res2Net-101& 72.2 & 89.4	 & 79.8	   & 68.9	 & 79.2     \\ \midrule   
Res2Net-v1b-50  & 72.2 & 89.5	 & 79.7	   & 68.5	 & 79.4     \\ 
Res2Net-v1b-101 & 73.0 & 89.5	 & 80.3	   & 69.5	 & 80.0     \\ \bottomrule   
    \end{tabular}
    \label{tab:keypoint}
\end{table}
}

\section{Conclusion and Future Work}

We present a simple yet efficient block, namely \ourM,
to further explore the multi-scale ability of CNNs at a more granular level.
The \ourM~exposes a new dimension, namely ``scale'',
which is an essential and more effective factor in addition to
existing dimensions of depth, width, and cardinality.
Our \ourM~module can be integrated with existing \sArt methods with no effort.
Image classification results on CIFAR-100 and ImageNet benchmarks
suggested that our new backbone network consistently performs favourably 
against its \sArt competitors,
including ResNet, ResNeXt, DLA, \etc.

Although the superiority of the proposed backbone model has been demonstrated
in the context of several representative computer vision tasks,
including class activation mapping, object detection,
and salient object detection,
we believe multi-scale representation is essential for a much wider range of
application areas.
To encourage future works to leverage the strong multi-scale ability of
the \ourM,
the source code is available on \url{https://mmcheng.net/res2net/}.

\section*{Acknowledgments}
This research was supported by NSFC (NO. 61620106008, 61572264),
the national youth talent support program, and
Tianjin Natural Science Foundation (17JCJQJC43700, 18ZXZNGX00110).

{\small
\bibliographystyle{ieee}
\bibliography{ref}
}

\newcommand{\AddPhoto}[1]{{\includegraphics[width=1in,height=1.25in,%
clip,keepaspectratio]{Authors/#1}}}

\ifdefined \GramaCheck
  \newcommand{\AuthorBio}[3]{#2 #3}
\else
  \newcommand{\AuthorBio}[3]{\vspace{-.2in}\begin{IEEEbiography}[\AddPhoto{#1}]{#2}#3\end{IEEEbiography}}
\fi

\AuthorBio{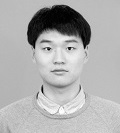}{Shang-Hua Gao}{
is a master student in Media Computing Lab at Nankai University.
He is supervised via Prof. Ming-Ming Cheng.
His research interests include computer vision,  machine learning, 
and radio vortex wireless communications.
}

\AuthorBio{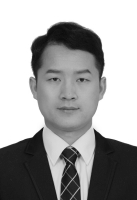}{Ming-Ming Cheng}{
received his PhD degree from Tsinghua University in 2012,
and then worked with Prof. Philip Torr in Oxford for 2 years.
He is now a professor at Nankai University, leading the
Media Computing Lab.
His research interests includes computer vision and computer graphics.
He received awards including ACM China Rising Star Award,
IBM Global SUR Award, \etc.
He is a senior member of the IEEE and on the editorial boards of IEEE TIP.
}

\AuthorBio{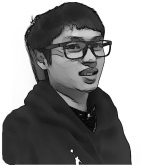}{Kai Zhao}{
Kai Zhao is currently a Ph.D candidate with college of computer science, Nankai University, 
under the supervision of Prof Ming-Ming Cheng. 
His research interests mainly focus on statistical learning and computer vision.
}

\AuthorBio{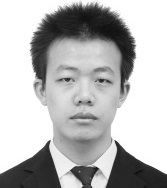}{Xin-Yu Zhang}{
is an undergraduate student from School of Mathematical 
Sciences at Nankai University. 
His research interests include computer vision and deep learning.
}

\AuthorBio{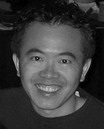}{Ming-Hsuan Yang}{
is a professor in Electrical
Engineering and Computer Science at University of California, Merced. 
He received the PhD degree in Computer Science from the University
of Illinois at Urbana-Champaign in 2000. Yang
has served as an associate editor of the IEEE TPAMI,
IJCV, CVIU, \etc.
He received the
NSF CAREER award in 2012 and the Google Faculty Award in 2009.
}

\AuthorBio{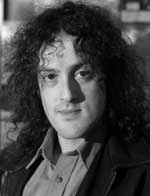}{Philip Torr}{
received the PhD degree from Oxford University. 
After working for another three years at Oxford, 
he worked for six years for Microsoft Research, first in Redmond, 
then in Cambridge, founding the vision side of the Machine
Learning and Perception Group. 
He is now a professor at Oxford University. 
He has won awards from top vision conferences, 
including ICCV, CVPR, ECCV, NIPS and BMVC. 
He is a senior member of the IEEE and a Royal Society
Wolfson Research Merit Award holder.
}

\vfill

\end{document}